\documentclass[journal]{IEEEtran}
\usepackage{graphicx}
\usepackage{todonotes}
\usepackage[utf8]{inputenc}
\usepackage{amsmath,amssymb,amsfonts}
\usepackage{textgreek}
\usepackage{graphicx}
\usepackage{color,bm}
\usepackage{wasysym}
\usepackage{tikz}
\usepackage{cite}
\usepackage{array,multirow}
\usepackage{booktabs} 
\usepackage{float}
\usepackage{balance}
\usepackage{hyperref}
\usepackage{mathtools, nccmath}
\usepackage{tabularx}
\usepackage{ragged2e}
\usepackage{subfigure}
\usepackage{threeparttable}
\usepackage{algorithm}
\usepackage{algpseudocode}
\usepackage{amsmath}
\algnewcommand{\Input}[1]{%
  \State \textbf{Input:} #1}
\algnewcommand{\Output}[1]{%
  \State \textbf{Output:} #1}
  
\DeclareGraphicsExtensions{.tif}
\begin{document}
\title{Deep Activity Model: A Generative Deep Learning Approach for Human Mobility Pattern Synthesis}

\author{Xishun Liao\textsuperscript{1}, Qinhua Jiang\textsuperscript{1}, Brian Yueshuai He\textsuperscript{2}, Yifan Liu\textsuperscript{1}, Chenchen Kuai\textsuperscript{3}, Jiaqi Ma\textsuperscript{1}\IEEEauthorrefmark{1}

\thanks{1 Xishun Liao, Qinhua Jiang, Yifan Liu, and Jiaqi Ma are with UCLA Mobility Lab at University of Los Angeles, Los Angeles, CA 90095, USA. \IEEEauthorrefmark{1}Corresponding author: Jiaqi Ma, jiaqima@ucla.edu }
\thanks{2 Brian Yueshuai He is with Civil and Environmental Engineering Department at University of Louisville, Louisville, Kentucky, 40208, USA}
\thanks{3 Chenchen Kuai is with Civil and Environmental Engineering Department at Texas A\&M University, College Station, Texas, 77840, USA}
}

\maketitle
\begin{abstract}
Human mobility plays a crucial role in transportation, urban planning, and public health, but current approaches face important limitations. Existing deep learning models tend to overlook the semantic interdependencies among activities and households and rely on restricted GPS data, while activity-based models depend on rigid assumptions and extensive data, making them costly and difficult to adapt to new regions, especially those with limited conventional travel data. To address these limitations, we propose a generative Transformer model that uses socio-demographic and household attributes to synthesize daily activity chains, with a location module assigning spatial zones to produce complete daily trajectories. Trained on open-source and widely available household travel survey data and then fine-tuned with local data, the model captures national activity patterns and transfers effectively to California, Washington, and Mexico City.

This approach offers significant potential for advancing synthetic human mobility modeling and provides urban planners and policymakers with improved tools for simulating transportation systems and supporting decisions in urban development and public health. Its practical utility is demonstrated through large-scale traffic simulations in Los Angeles County. Compared to the SCAG Activity-Based Model (ABM), the proposed method produces consistent spatial demand patterns, achieving an activity location cosine similarity of 0.997 and a network-level vehicle-miles-traveled Mean Absolute Percentage Error (MAPE) of 4.97\%. Compared to real-world observations from Caltrans PeMS, the simulated traffic achieves MAPE of 5.85\% for traffic volume and 4.36\% for speed on California’s I-405 corridor. This demonstrates the practicality of learning-based synthetic mobility for regional simulation and planning.

\end{abstract}


\section{Introduction}\label{IntroSection}
\subsection{Motivation}
\IEEEPARstart{U}nderstanding and synthesizing human mobility patterns has gained considerable importance as population growth, increasingly complex travel behaviors and diverse societal needs reshape modern transportation systems \cite{ma2024mobility}. Human mobility impacts numerous facets of contemporary life, including traffic management, air quality, energy consumption, and public health. The COVID-19 pandemic, for instance, substantially altered mobility behaviors due to widespread shifts to remote work and decreased travel demand~\cite{wang2021impact,zhang2021effect, 11423811, hoppe2023improving, kraemer2020effect}. Additional factors affecting human mobility include traffic congestion and safety~\cite{wu2019agent,tselentis2023driver, liao2023realaccident}, citizen well-being~\cite{pappalardo2016analytical}, events~\cite{abidi2023mobility}, air pollution~\cite{bohm2021quantifying}, and energy and water consumption~\cite{mohammadi2017urban}. Consequently, modeling human mobility has garnered substantial research interest due to its relevance in addressing these critical issues.

Traditional models of human mobility include Activity-Based Models (ABMs), developed in the late 1990s~\cite{ben1998activity}. ABMs simulate the sequence of individual activities and predict interdependent activity choices, making them popular with Metropolitan Planning Organizations in the U.S.~\cite{bradley2008design}. For instance, ABM is used to predict activity patterns and travel demand for Southern California, a region with about 26 million population~\cite{he2022connected,jiang2022connected}. However, ABMs have significant limitations: data collection, model development, and calibration are costly and time-consuming; their complexity leads to high computational demands; and they rely on numerous assumptions, limiting their adaptability to other regions.

Data-driven approaches, including deep learning (DL) and generative algorithms, offer promising alternatives to address the limitations of ABMs in replicating human mobility trajectories. These methods can accurately mimic real-world patterns, aligning closely with observed data~\cite{karamshuk2011human,zhong2023estimating,pappalardo2018data,liao2024reconstructing,zheng2023fairness,11423529}. Prior research has focused on spatial (e.g., travel distance~\cite{gonzalez2008understanding}, preferred locations~\cite{pappalardo2015returners}) and temporal attributes (e.g., activity schedules~\cite{rinzivillo2014purpose}). Additionally, over 90\% of human mobility patterns exhibit recurring motifs, indicating that pre-trained models can be adapted to various regions~\cite{schneider2013unravelling,cao2019characterizing}.

Despite their advantages, current DL models often rely on specialized mobility data sources like GPS, social media, and communication records~\cite{fontes2023leveraging, van2023engaging, gonzalez2008understanding, ma2025learning}. These sources differ in format and availability, making cross-dataset and regional adaptation challenging. Additionally, access to such data can be costly, limited, and difficult, reducing model flexibility. Many models focus solely on spatial-temporal trajectories, overlooking the semantic relationships between activities and socio-demographic factors. These limitations highlight the need for adaptable models that can capture the complex interdependencies of activities, locations, and personal attributes.

\subsection{Objective and Study Scope}

In this paper, we aim to address the limitations of existing ABMs and data-driven mobility methods by introducing a novel generative deep learning model, referred to as the Deep Activity model. Our primary objective is to synthesize human mobility patterns only based on agent profiles and to reveal fundamental and generic mobility trends within a given region. We interpret human mobility patterns as chains of activities (i.e., travel demand) and locations that individuals visit, influenced by socio-demographic factors and environmental conditions. Unlike previous approaches that focus on predicting the next point of interest (POI) or activity based on historical data during inference, our model emphasizes generation, producing synthetic and realistic human mobility patterns based solely on input socio-demographic profiles. Once trained, it can generate high-fidelity activity chains for new individuals or households without requiring any prior trip history. This makes it suitable for data-sparse regions or the simulation of hypothetical populations, as it does not depend on observed historical trajectories during generation.

\begin{figure}
  \centering
  \subfigure[Household travel survey data]{

    \includegraphics[width=0.8\linewidth]{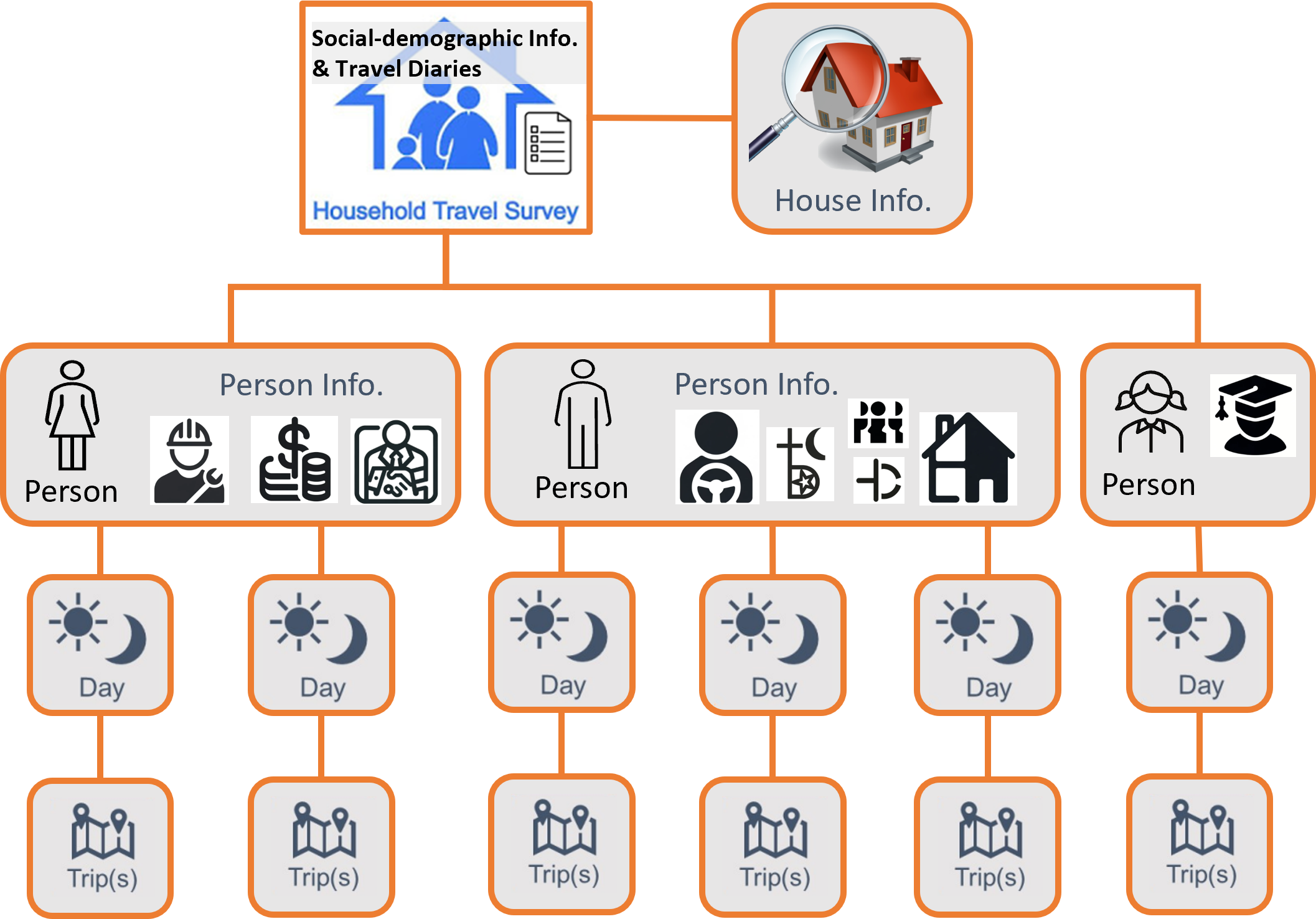}
    }
  \hfill
  \subfigure[Activity chains examples]{
    \includegraphics[width=0.7\linewidth]{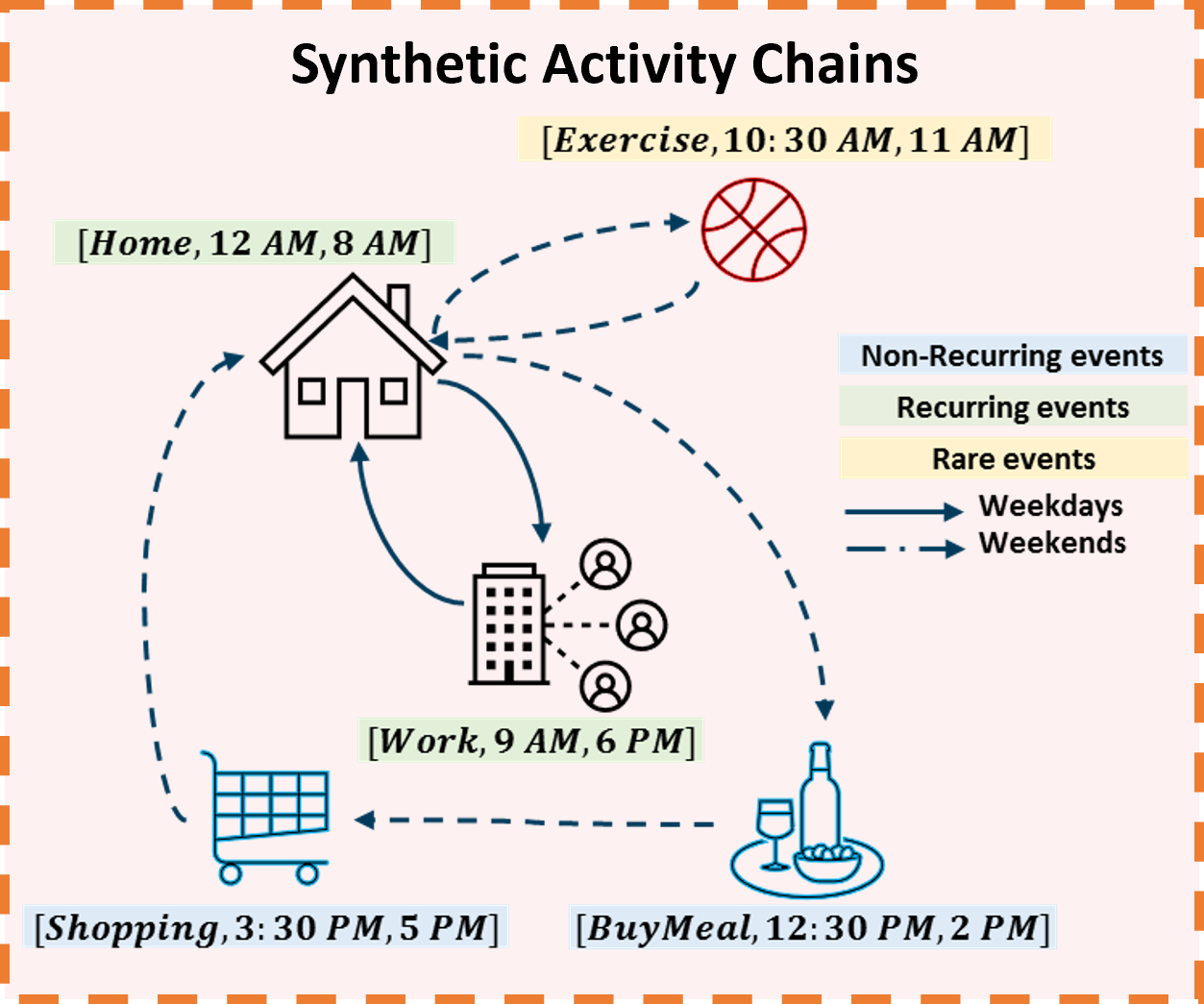}
    }
  \caption{Model human mobility pattern using HTS data. (a) HTS data includes information about each household member's socio-demographics, household characteristics, and daily non-commercial travel across all modes, including details about travelers, their households, and vehicles~\cite{mcguckin2018summary}. (b) Typical weekday and weekend activity chains in HTS.}
  \label{fig:intro}
  \vspace{-3mm}
\end{figure}

The scope of this research is centered on modeling human mobility at both the demand and trajectory levels. Our Deep Activity model is derived from trip diaries available in household travel survey (HTS) data (Fig.\ref{fig:intro} (a)). We utilize the concept of an "activity chain" (Fig.\ref{fig:intro} (b)), which represents a one-day sequence of activities for individuals. Leveraging regional population information, the Deep Activity model generates realistic and varied activity chains. With assigning specific locations to each activity in the chain, our proposed method effectively captures the underlying human mobility patterns of the target region. Furthermore, the model can be fine-tuned for specific regions, capturing unique patterns in California, the Puget Sound region (Washington state), and Mexico City.
\subsection{Contributions}
With such synthesis capability, the automatic generation of transportation system simulation models becomes much more feasible. This addresses the significant challenge of the extremely high costs and labor-intensive nature of hand-crafted models that have persisted for a long time. More implications include facilitating location choice models and urban planning models by integrating the Deep Activity model. This paper sets the foundation for human mobility synthesis, enabling the automatic generation of data for new regions and significantly advancing the field of transportation modeling. Our key contributions include:

\begin{itemize}
    \item We advance the study of human mobility pattern synthesis by introducing a learning-based generative framework that integrates travel demand generation and travel trajectory creation while incorporating regional mobility characteristics. Our approach ensures privacy by relying on aggregated survey data rather than individual-level trajectory data. Once the model is trained, it can generate realistic activity chains without requiring historical trajectories. By introducing randomness during inference or simulating synthetic populations, the model can produce unlimited plausible activity chains for the target region. Additionally, we introduce new performance metrics for effective evaluation.
    
    \item We develop a deep learning model that generates synthetic human mobility data based solely on socio-demographic and household characteristics. The model is transferable across regions and can be fine-tuned with limited local data, making it especially useful in data-sparse settings. By leveraging HTS data, optimizing input representation, and designing an effective loss function, even a standard Transformer architecture achieves strong performance.
    
    \item We explicitly model the interdependencies among activities of household members, investigating how decisions made by one member influence others, thus capturing household-level mobility dynamics.
   
    \item We explore a standard technique for multivariate, multi-objective data balancing to process the ubiquitous HTS data. This pioneering approach enhances the applicability of HTS data in deep learning models for human mobility studies, providing significant benefits to the transportation modeling and planning community.
\end{itemize}

\section{Related work}
Human mobility has gained attention recently due to increased data availability, computing advancements, and AI techniques. Research in this field focuses on two main tasks: generation and prediction~\cite{luca2021survey}. Generative models aim to create realistic human trajectories, replicating real-world mobility flows, while prediction models forecast individual movements or crowd flows based on historical data~\cite{zhou2021selflocation1,comito2020nextlocation2,feng2018deepmovehistory2}. This paper focuses on generative models, specifically addressing the challenge of generating realistic activity chains in human mobility patterns.

\textbf{Transportation Models}. ABMs are the state-of-the-practice model to generate human activity patterns. An ABM is a type of modeling approach used in transportation planning and urban studies to predict and analyze individual activity patterns and travel behavior. ABMs aim to understand and simulate how people make decisions about their daily activities, such as work, shopping, education, recreation, and other social and personal activities, and how these activities influence their travel choices and travel patterns. Bowman and Ben-Akiva~\cite{bowman2001activity} proposed an ABM prototype to predict individual activity and travel schedule based on discrete choice models and forecasted the travel demand of the Boston metropolitan area. Goulias et al.~\cite{goulias2011simulator} developed SimAGENT, an ABM to simulate activity and travel patterns in Southern California. SimAGENT comprises five components: a population synthesizer, socioeconomics micro-simulator, land use and transportation systems, daily activity-travel micro-simulator, and transportation simulation~\cite{bhat2012household}. However, the development of ABMs requires extensive data collection, making them expensive and resource-intensive. Additionally, their reliance on localized data limits their adaptability to different regions, which restricts their broader applicability.

\textbf{Model-Based Data Driven Methods}. Data-driven models offer an alternative approach for activity generation problems. The Exploration and Preferential Return (EPR) model, a stochastic method, simulates human mobility by balancing exploration of new locations with returns to previously visited ones~\cite{song2010modelling}.  In the EPR model, individuals move through a spatial environment, making decisions influenced by location popularity and distance. The model considers the balance between exploration and preferential return and evolves over time. Enhancements like TimeGeo by Jiang et al.~\cite{jiang2016timegeo} add temporal choices, such as home-based tour number, dwell rate, and burst rate, along with a hierarchical multiplicative cascade method to measure generated trips and land use. These improvements bypass HTS data limitations by offering a flexible, data-driven framework. However, model-based data-driven methods often require prior expert knowledge, and their simple implementation mechanisms may constrain realism~\cite{NHTSCA2019,agriesti2023bayesian}.

\textbf{Deep Learning Approaches}. Multiple DL models have been used to model human mobility and generate human activities and trajectories, including fully connected networks, recurrent neural networks (RNNs), attention mechanisms, convolutional neural networks (CNNs), and generative models such as Generative Adversarial Networks (GANs) and Diffusion Models. For a detailed review, see Luca et al.~\cite{luca2021survey}. The limitations of model-based, data-driven methods can be addressed by generative models, as they can simultaneously incorporate various aspects of human trajectories (e.g., spatial and temporal features) and capture complex, non-linear relationships in the data. GANs and diffusion models, in particular, are well-suited for modeling complex data structures like detailed trajectory data due to their ability to learn intricate movement patterns. However, these models require large amounts of high-quality data to train effectively \cite{kumar2023navigating}, which poses challenges when dealing with simpler data formats, such as survey data, where individuals have fewer activities and locations. However, the performance of DL models heavily depends on the quantity and quality of data, and they are constrained by the fact that human mobility data is often expensive or difficult to access (e.g., requiring Non-Disclosure Agreements). Thus, exploring the potential of DL models to synthesize human mobility patterns using ubiquitous and open-source data is essential.

\textbf{Large Language Model Approaches}. Recent advances in Large Language Models (LLMs) have led to increased research interest in applying these models to human mobility pattern analysis and generation. LLMs offer significant advantages for mobility modeling due to their ability to process contextual information and generate complex sequences with minimal domain-specific training requirements. Several recent studies have explored LLM applications in this domain. TSI-LLM enhances semantic inference from trajectory data through spatiotemporal formatting and context-inclusive prompts \cite{luo2024decipheringhumanmobilityinferring}. Mobility-LLM interprets check-in sequences using a visiting intention memory network combined with travel preference prompts \cite{gong2024mobilityllmlearningvisitingintentions}. TrajLLM proposes a modular LLM-empowered, agent-based framework that integrates persona generation and activity selection to simulate realistic human trajectories \cite{ju2025trajllmmodularllmenhancedagentbased}. For activity chain generation specifically, a retrieval-augmented generation (RAG) approach has been developed \cite{Liu2024HumanMobility} that integrates public statistical and socio-demographic data for generating realistic activity sequences. This represents the most sophisticated state-of-the-art algorithm for activity chain generation tasks.

\begin{figure*}
  \centering
  \includegraphics[width=0.99\textwidth]{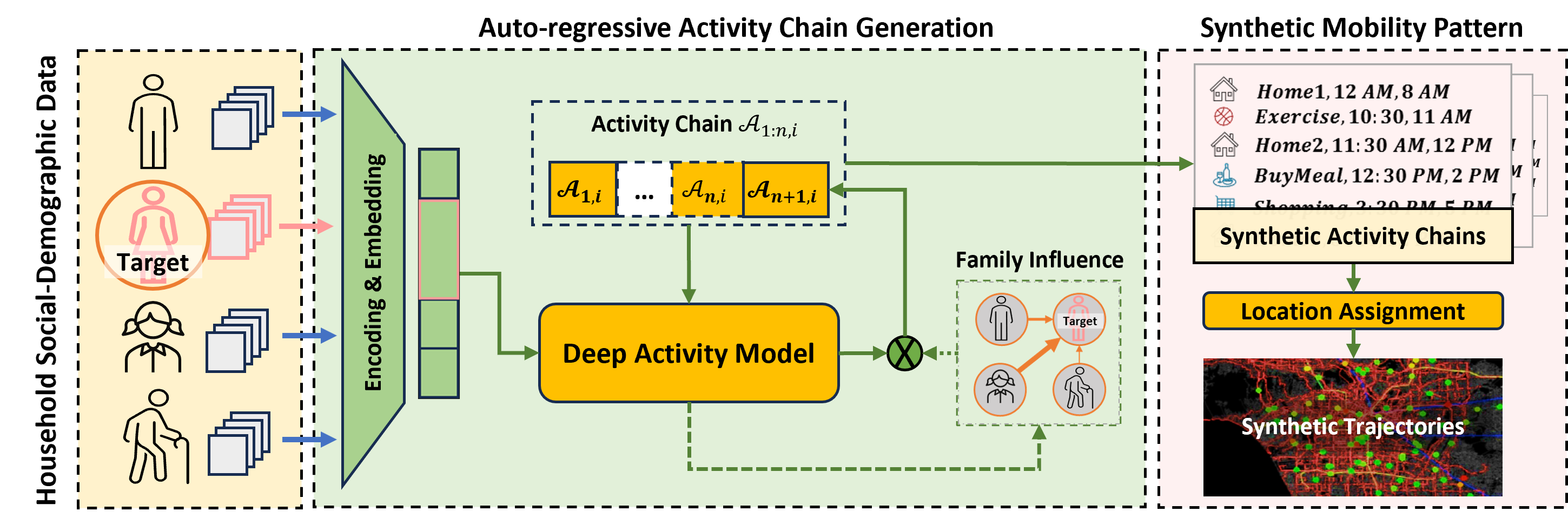}
  \caption{Workflow of activity chain generation. Given the synthetic socio-demographic information and household characteristics of each agent, the model auto-regressively synthesizes the agent's activity chain and the location of each activity.}
  \label{fig:system_workflow}
  \vspace{-5mm}
\end{figure*}

\section{Problem Formulation}
One of the fundamental principles of ABM is that "the demand for travel is derived from the demand for activities"~\cite{bowman2001activity}. This principle highlights the nature of human mobility, where activities form the foundation of human trajectories. In this study, the concept of an "activity chain" is used to describe the structure of these trajectories, and the human mobility of a region can be represented by the activity chains of its population.

We denote $i$ for an agent. An activity chain for the agent $i$ is a time-ordered sequence $A_i=\left\{A_{1, i}, A_{2, i}, \ldots, A_{j, i}\right\}$, where $A_{j, i}=\left[T_{j, i}, S_{j, i}, E_{j, i}\right]$ represents the $j$-th activity conducted by agent $i$. Here, $T_{j,i}$ is the activity type of $A_{j, i}$. $S_{j, i}$ and $E_{j, i}$ stand for start time and end time of $A_{j, i}$, respectively. Then the mobility trajectory can be expressed as $Traj_i = \left\{ \left( A_{1, i}, Z_{1,i} \right), \ldots, \left( A_{j, i}, Z_{j,i} \right)  \right\}$, where $Z_{j,i}$ denotes the zone-level location where the $A_{j, i}$ occurs. A generative model $M$ can generate activity chains $A_i$ for each individual $i$, given socio-demographic attributes of the target agent and other household members, $D_{k, i}=\left\{d_{1, i}, d_{2, i}, \ldots, d_{K, i}\right\}$, where $d_{K, i}$ represents the $k$-th socio-demographic attributes.

\section{Dataset}
\subsection{Household Travel Survey}
To generate the activity chain for each individual in the region of interest, we rely on data collected through the HTS, following a standard format across different regions, as introduced in Fig. \ref{fig:intro}(a). In the US, this survey is usually conducted by federal agencies or state agencies to gather detailed information about people’s travel behaviors. The Federal Highway Administration (FHWA) administered the National Household Travel Survey (NHTS) for the United States~\cite{federal_highway_administration_2022}. To uncover regional activity patterns, many states also conducted statewide HTS~\cite{new_york_metropolitan_transportation_council_2012,calidata,pugetdata,mexicodata}. This travel-diary data source is also widely available in many countries as government agencies need such data for various purposes of public resource management. Unlike trajectory data, which includes detailed location traces that could potentially identify individuals when combined with demographic information, HTS data focuses on summarized activity patterns and aggregated travel behaviors, which helps to mitigate privacy risks.

Building on this standardized HTS data, the Deep Activity model can be easily trained and transferred to other regions. In this study, the generic model is developed using the 2017 NHTS to enhance adaptability across regions, leveraging its large dataset of over 129,600 US households, which includes demographics, activity patterns, and travel behaviors for each household member. As presented in TABLE ~\ref{table:activity_type_table}, the activity types in NHTS are aggregated to 15 types based on the locations of activities. For instance, regular home activities and work from home are grouped as the home activity. Besides NHTS, the 2010–2012 California Household Travel Survey \cite{calidata} (collected from 42,500 households), the 2017 Puget Sound Regional Travel Study \cite{pugetdata} (collected from 3,285 households), and the 2017 Origin-Destination Survey across the Mexico City Metropolitan Area \cite{mexicodata} (collected from 66,625 housing units) are adopted in this study for transferability exploration.

\begin{table}[ht]
    \centering
    \caption{Activity category code and their corresponding descriptions from 2017 NHTS}
    \begin{tabular}{|c|c|c|c|c|c|}
        \hline
        1 & Home & 2 & Work  & 3 & School \\ \hline
        4 & Care giving & 5 & Buy goods & 6 & Buy services \\ \hline
        7 & Buy meals & 8 & General errands & 9 & Recreational \\ \hline
        10 & Exercise  & 11 & Visit friends & 12 & Health care \\ \hline
        13 & Religious & 14 & Something else & 15 & Drop off/Pick up\\ \hline
    \end{tabular}
    \label{table:activity_type_table}
\end{table}
\subsection{SCAG ABM Data}

In addition to the travel survey data, we utilize synthetic mobility data, specifically the SCAG ABM data~\cite{scag2020}, for this study. The SCAG ABM dataset represents simulated human mobility patterns, providing detailed synthetic single-day activity diaries across six counties in Southern California, encompassing a population of over 19 million. The model captures 24-hour travel demand patterns at a 15-minute temporal resolution, including start and end times, types of activities, and zonal-level locations for each individual agent.

One notable advantage of the SCAG ABM data is the inclusion of zonal-level location data for each simulated activity. These zones, known as Transportation Analysis Zones (TAZs), contain demographic and spatial information about the residents and destinations within each zone, serving as both origins and destinations of trips. The use of this data significantly enhances our ability to validate the spatial-temporal performance of the mobility patterns generated by the model proposed in this study, providing a robust framework for further analysis and validation. In this study, we select Los Angeles (LA) County, which contains 5,967 TAZs, as the target area for validation.

\subsection{Data Preparation}

The first step in training the model involves preparing the agent information and activity chain pairs, which includes selecting relevant features for the agent information and encoding the activity chain data.

According to ABM, \textbf{socioeconomic and demographic attributes} significantly influence individual activity patterns and travel choices. In the context of the activity generation model, the function of these attributes is similar to a prompt (as in the language model), determining the start of activity sequence generation and influencing the entire sequence generation process. These attributes include typical individual characteristics, such as gender, race, age, employment status, and job category, capturing personal and professional demographics. Education level and student status provide insights into the academic background and current academic involvement. 

\textbf{Household-related} attributes such as the number of persons, relationships within the household, home ownership, and household size are also considered for their impact on daily routines and mobility. The number of vehicles owned, workers in the household, and the household employed count are indicative of transportation needs and capabilities. Household income level, along with the percentage of renter-occupied housing in the household's location, offers a socioeconomic perspective. Finally, the household life cycle stage captures adult composition, presence of children, and retirement status, reflecting evolving mobility needs. For example, a household with two adults and a child aged 0–5 indicates caregiving responsibilities that shape daily travel.

Additionally, \textbf{zonal attributes}, including the population density, housing units, and the classification of the residence type as rural or urban, provide a geographical context. These demographic features collectively offer a comprehensive description of an individual's background. Finally, there are 13 personal attributes, 13 household shared attributes, in total 26 attributes selected to describe one individual. In conclusion, the personal and household features used for the model are presented in Table ~\ref{table:nhtsfeature}.

The attribute data, originally in text or label format, is transformed into categorical data. Not everyone answers all the privacy-related questions in the survey, so for any attributes left blank or marked as 'not responded' in the NHTS, we used a dummy number for encoding. The continuous activity start and end times are encoded with segmenting a 24-hour day into 96 intervals, each lasting 15 minutes, and numerically encoded from 1 to 96 to represent the time slots. 

Regarding the \textbf{SCAG ABM dataset}, we sample subsets from the overall population of 10 million individuals for the purposes of model transfer learning and validation. Specifically, we prepare two subsets: a smaller sample of 100,000 individuals, which is used for model transfer learning, and a larger sample of approximately 1 million individuals, employed to assess the scalability of the model trained on the smaller subset to a larger population. The activity types in the SCAG ABM data are mapped to correspond to 15 categories consistent with the NHTS data, and the time variables are similarly encoded into 96 time slots, following the same procedure used for the NHTS data.

\begin{table}[h]
\centering
\caption{Socio-demographic attributes from HTS dataset}
\label{table:nhtsfeature}
\begin{tabular}{ll}
\toprule
\textbf{Attribute Name}                        &                           \\
\midrule
Driver's License Status                       & Number of Workdays                              \\
Education Level                              & Job Category                       \\
Gender                                        &  Age                               \\
Racial/Ethnic Identity                        & Weekly Transit Usage                               \\
Household Role                                & Household Income                            \\
Current School Grade Level                    & Household Size                                  \\
Employment Location Type                      & Household Vehicles                    \\
Number of Jobs                                & Home Ownership Status                           \\
Employment Status                             & Household Students                  \\
Household Licensed Drivers                    & Household Life Cycle Stage                         \\
Household Employed Members                    & Type of Residence    
\\
Housing Status                                & Housing Density    
\\
Population Density                    & Renter-Occupied Housing Ratio    
\\
\bottomrule
\end{tabular}
\vspace{-2mm}

\end{table}

\section{Methodology}\label{MethodSection}

The workflow for generating activity chains for an individual using our Deep Activity model is illustrated in Fig. \ref{fig:system_workflow}. The process begins with household socio-demographic data, which includes information about the target person and any family members. The Deep Activity model captures the influence of household members on the target person and ensures that the generated activity chain reflects real-world interdependencies. The model then auto-regressively generates a sequence of activities, forming the target person's activity chain $A_i$. Finally, location is assigned for each activity in $A_i$, completing the synthetic mobility trajectory generation.

\begin{figure*}
  \centering
  \includegraphics[width=0.99\textwidth]{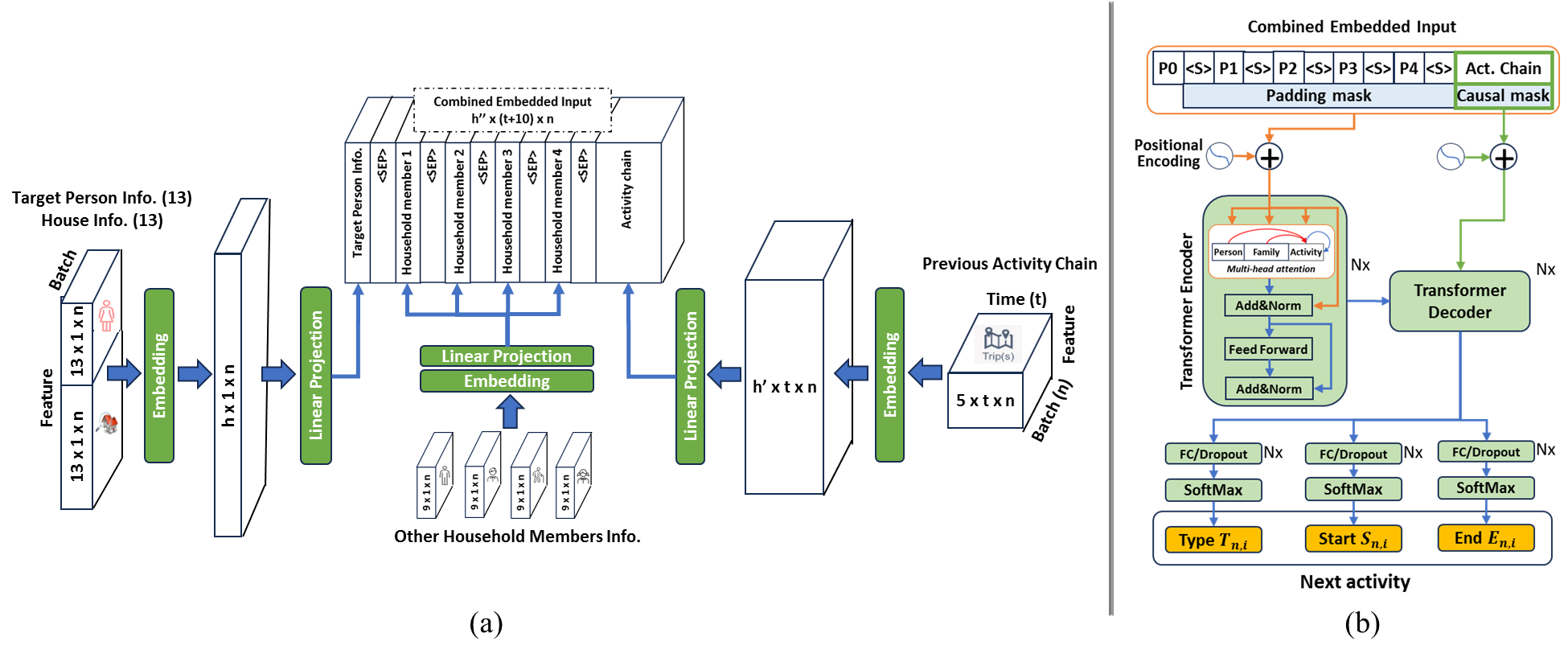}
  \caption{Deep Activity model architecture. (a) Input data construction. (b) Transformer-based network architecture with well-designed data injection.}
  \label{fig:system_structure}
  \vspace{-5mm}
\end{figure*}

\subsection{Model Architectures}

To generate an activity chain, comprising activity types alongside their corresponding start and end times, based on the demographic attributes of individuals. The structure of the activity chain generation problem is analogous to text generation tasks tackled by language models. Just as words in a sentence follow a logical sequence based on context, activities in a person's daily routine are sequentially dependent on preceding activities and time constraints. Hence, a model based on Transformer~\cite{vaswani2017attention} is developed and trained for the activity chain generation task, as shown in Fig. \ref{fig:system_structure}.

\textbf{Data structure design}. To analyze the influence of household members and previous activities on the target person's decision, we developed an innovative data concatenation strategy. This approach combines embedded social-demographic data, data of other household members, and embedded activity data within the time domain. To standardize dimensions across these diverse data sets, we integrated fully connected layers and employed learnable delimiters $<$SEP$>$ for separation. Fig. \ref{fig:system_structure}(a) illustrates this process for a household with five members, demonstrating how the data is transformed into a comprehensive feature vector. This vector subsequently serves as input for the network's deeper layers, enabling more nuanced analysis. Our model is designed to accommodate households of up to five members, a decision informed by statistical analysis of the NHTS dataset, which reveals that 95\% of households do not exceed this size.

\textbf{Feature Embedding}. We employ embedding layers to map each categorical variable into a continuous space, utilizing the Embedding function~\cite{paszke2019pytorch}, which can be optimized through backpropagation. To ensure an accurate representation of categories, we created a distinct embedding layer for each categorical attribute. Formally, for a categorical feature $c$ with $N$ unique categories, the embedding function is defined as: $ E_c:{1,2, . ., N} \rightarrow R^d $, where $d$ represents the dimension of the embedding space for that feature. The optimal value of $d$ was determined through validation performance.

As illustrated in Fig. \ref{fig:system_structure} (a), the embedding layer processes the activity chain data, represented by a tensor with dimensions (5, t, n), where these dimensions correspond to features, time, and batch size, respectively. Additionally, another embedding layer handles the target individual's social-demographic data, shaped as (26, 1, n), and the data pertaining to the target person and other household members. These diverse data sets are then seamlessly combined, utilizing five $<$SEP$>$ delimiters to maintain a clear separation between different data types.

\textbf{Network structure design}. The network adopts a Transformer-based encoder-decoder architecture. As shown in Fig.\ref{fig:system_structure}(b), the encoder receives combined embeddings of personal, household, and prior activity information. Padding and causal masks are applied to control attention during training. Padding masks in the encoder exclude positions corresponding to nonexistent household members and special tokens such as <SEP>, ensuring focus on valid inputs. Causal masks in the decoder preserve the autoregressive nature of sequence prediction by blocking attention to future tokens, maintaining left-to-right generation during both training and inference. The decoder then takes the combined activity sequence and the output (memory) from encoder as additional context. By processing the entire context in the encoder and focusing on the next activity prediction in the decoder, the model captures complex dependencies and interactions. Unlike traditional autoregressive models such as RNNs, this design supports parallel processing and more effectively captures dependencies within all inputs. It simultaneously accounts for personal, household, and prior activity influences on an individual's next activity decision, enhancing prediction accuracy and providing deeper insights into the drivers of activity choices. Next, positional encoding are added to both encoder and decoder inputs to retain temporal information. Finally, the model generates predictions for the activity type $T_{j,i}$, start time $S_{j,i}$, and end time $E_{j,i}$, forming the next activity. The prediction process continues until either the activity marked as the end-of-the-sentence (EOS) is predicted, or the chain reaches the maximum length, at which point the prediction terminates. 

\subsection{Loss functions}
Predicting activity type is a classification task. Because the day was segmented into 96 intervals, each lasting 15 minutes, as described in Appendix B, predicting start and end times is also a classification task.  \textbf{Cross-entropy loss}, $\mathbf{L_{CE}(y, \hat{y})} = -\sum_j y_j \log(P(\hat{y}_j))$, is commonly used to measure the discrepancy between predicted probabilities and actual outcomes and is the first loss term to minimize activity mismatch. However, given the uncertainty of human activity, time prediction should not be overly strict but rather should allow for a certain level of deviation. Therefore, we incorporate a custom loss function that includes soft labels to allow the prediction results to deviate within a small window, enhancing the flexibility and robustness for start and end time prediction.

\textbf{Soft label loss $\mathbf{L_s}$}. To introduce tolerance in time predictions, we use soft label loss, an enhancement of cross-entropy loss that allows multiple classes to carry non-zero probability. Instead of assigning a one-hot label (e.g., 100\% weight to a single 15-minute time slot), we assign a high weight to the true class and smaller weights to adjacent time slots within a predefined window. This forms a probability distribution around the ground truth time. This allows flexibility while preserving structure. Formally, for a given true label $y_k$, we define a soft label vector $SL \in \mathbb{R}^K$ as:
\begin{equation}
SL_{k} = 
\begin{cases}
w_m & \text{if } k = y_i \\
w_s & \text{if } 1 \leq |k - y_i| \leq n_s \\
0 & \text{otherwise}
\end{cases}
\end{equation}

\noindent where $i = 1, \dots, N$ indexes the training samples in a batch; $k = 1, \dots, K$ indexes the time bins (e.g., 96 bins for a day divided into 15-minute intervals); $w_m$ is the main class weight (e.g., 1.0); $w_s$ is the side weight for neighboring classes (e.g., 0.1); $n_s$ is the number of side steps included in the soft label window.

Then the soft label cross entropy loss is computed as:
\begin{equation}
L_s = -\frac{1}{N} \sum_{i=1}^{N} \sum_{k=1}^{K} SL_{i,k} \log (P_{i,k} + \epsilon)
\end{equation}
\noindent where $\epsilon$ is a small constant added to prevent the logarithm of zero.

Additionally, to guarantee that the generated sequence of activities adheres to a logical chronological order, two specialized time penalty losses, i.e., \textbf{temporal order loss $(\mathbf{L_o})$} and \textbf{sequential timing loss $(\mathbf{L_{seq}})$}, are incorporated. These losses ensure that the predicted end time of an activity does not precede its start time and that the end time of a preceding activity does not exceed the start time of the subsequent activity.

\begin{equation}
\begin{gathered}
L_o=\frac{1}{N} \sum_{i=1}^N \max \left(0, t_{i,t-1}^{\text {end }}-t_{i,t}^{\text {start }}\right),   \\
L_{seq}=\frac{1}{N} \sum_{i=1}^N \max \left(0, t_{i,t}^{\text {start }}-t_{i,t}^{\text {end }}\right)
\end{gathered}
\end{equation}

\noindent where $t_{i,t-1}^{\text{end}}$ denotes activity end time at $t-1$ step, and $t_{i,t}^{\text{start}}$ means activity start time at $t$ step.

Our final loss $L$ combines five loss terms as below:

\begin{align}
L &= w_1 \cdot L_{CE}(T, \widehat{T}) + w_2 \cdot L_s(S, \hat{S}) \nonumber \\
  &\quad + w_3 \cdot L_s(E, \hat{E}) + w_4 \cdot L_o + w_5 \cdot L_{seq}
\end{align}

\subsection{Data Balancing}

To create a fair and representative training dataset for model development, data balancing is essential to address the imbalances in HTS data. Most people follow similar activity patterns ~\cite{schneider2013unravelling,cao2019characterizing}, which can overshadow less common activities and lead to biased models favoring the majority class. To address this, we developed a multivariate, multi-objective data balancing technique for curating the HTS dataset. In this study, three target features need to be balanced, as shown in Table \ref{tab:target_features_example}, which showcases examples of these target features, with each data sample representing a day's activities for an individual.

\begin{algorithm}
\small
\caption{Data balancing for multiple target features}
\footnotesize
\begin{algorithmic}
\Input (1) Training dataset, (2) Adjustment rate at each step: $step_{size}$
\Output Weights $W$ assigned to training dataset for data resampling
\begin{algorithmic}[1] 
\State Select $n$ target representation features features and calculate the original distributions: $D_{ori} = \{O_1, O_2, ..., O_n\}$
\State Ideal distribution for $n$ target features: $D_{ideal} = \{I_1, I_2, ..., I_n\}$
\State Initialize target distributions: $D_{tar} = \{T_1, T_2, ..., T_n\} = D_{ori} = \{O_1, O_2, ..., O_n\}$
\Repeat
    \For{each feature $i$ in $n$ target features}
        \State // Calculate the differences between adjusted and target share for each class in feature $i$:
        \State $D_i = T_i - I_i$
        \State \Comment{Adjust the elements based on its difference and the} $step\_size$:
        \State $T_i = T_i - D_i \cdot step\_size$
        \State // Ensure the sum of percentages remains equal to 1 by normalizing the values:
        \State $F_i = 1/sum(T_i)$ \Comment{define a normalization factor}
        \State $T_i = T_i \cdot F_i$ \Comment{update the target distribution}
        \State // Calculate weights for each sample ($W$) using the raking algorithm:
        \State $W = raking(D_{ori}, D_{tar}, train\_data)$
    \EndFor
\Until{raking algorithm not converging or $|D_{tar} - D_{adj}| < threshold$}
\State \Return $W$
\end{algorithmic}
\label{Algorithm1}
\end{algorithmic}
\end{algorithm}

\begin{table}[h!]
\centering
\large
\caption{Examples of target features to be balanced}
\resizebox{\columnwidth}{!}{ 
\begin{tabular}{|c|c|c|c|}
\hline
\multicolumn{1}{|c|}{\textbf{id}} & \multicolumn{1}{c|}{\textbf{Activity Type}} & \multicolumn{1}{c|}{\textbf{Chain Length}} & \multicolumn{1}{c|}{\textbf{Duration (15-min)}} \\ \hline
1 & \begin{tabular}[c]{@{}c@{}} \{Home, Work, Home,\\ Exercise, Home\}\end{tabular} & 5 & \begin{tabular}[c]{@{}c@{}}\{28, 32, 4, 7, 6\}\end{tabular} \\ \hline
2 & \begin{tabular}[c]{@{}c@{}}\{Home, School,\\ Buy meals, Home\}\end{tabular} & 4 & \begin{tabular}[c]{@{}c@{}}\{25, 35, 8, 24\}\end{tabular} \\ \hline
3 & Home, Work, Home & 3 & \{30, 60, 15\} \\ \hline
\end{tabular}
}
\label{tab:target_features_example}
\end{table}

The proposed data balancing method iteratively calculates weights for each data sample based on its significance, then performs random resampling with replacement to produce a balanced training dataset. This process can be summarized as in \textbf{Algorithm 1}, involving five key steps:

\textbf{Step 1. Feature representation}:
As indicated in TABLE \ref{tab:target_features_example}, activity type and duration are recorded in sequence, while the length of activity chain is a singular value, making it difficult to balance. Therefore, the most frequently occurring activity type ("mode type") and activity duration ("mode duration") are selected from each activity chain, as representations of the original target features, excluding the first and last home activities, since most of activity chains start and end at home.

\textbf{Step 2. Initial distribution $D_{ori}$ computation}: Calculate the real class distribution for each target feature. 

\textbf{Step 3. Target distribution $D_{tar}$ specification}: Set $D_{tar}$ as an intermediate between the actual and ideal distributions $D_{ideal}$ (uniform distribution) to facilitate convergence.

\textbf{Step 4. Sample weights calculation}: Compute sample weights to individual samples using the raking algorithm~\cite{deville1993generalized}, based on $D_{ori}$ and $D_{tar}$ from \textbf{Steps 2} and \textbf{3}.

\textbf{Step 5. Iterative refinement}:
If convergence is achieved, adjust $D_{tar}$ closer to $D_{ideal}$ and recalculate sample weights by repeating \textbf{Steps 3} and \textbf{4}.

As a result, the target distribution for each feature is set between the observed and uniform distributions, so that very common patterns are down weighted and rare activities are up weighted without flattening the empirical distribution. This increases training diversity while largely preserving the marginal distribution of total daily duration. The main side effect is a moderate oversampling of rare classes, which slightly increases variance in those bins but, as illustrated in Fig.\ref{fig:Databalancing} for Mexico City, reduces divergence in activity type and transition metrics without noticeably affecting temporal distributions.

\subsection{Model Transfer}
As aforementioned in Section \ref{IntroSection}, modeling human mobility patterns in regions with limited data is challenging using traditional approaches. Even with advanced deep learning methods, such as transformer models, the data-hungry nature of these models can limit their effectiveness when datasets are small~\cite{vaswani2017attention}. By leveraging the concept of transfer learning~\cite{zhuang2020comprehensive}, we can address this challenge effectively. The proposed Deep Activity model, initially trained on the NHTS dataset (160,000 training samples), serves as a generic pre-trained model, which can then be fine-tuned using the limited local HTS data, adapting the generic model to the specific characteristics of smaller regions. The fine-tuning the Deep Activity model involves three primary steps:

\textbf{Step 1: Adding new layers to adapt to new features}. Augment the existing architecture with new layers to handle region-specific features and complexities, enabling better representation of diverse activity patterns.

\textbf{Step 2: Freezing part of the pre-trained model}. Initially freeze certain parts of the pre-trained NHTS model to maintain stability and leverage learned representations, preventing overfitting and ensuring effective capture of regional features.

\textbf{Step 3: Fine-tuning with regional data and unfreezing layers}. Train the modified model using regional datasets, updating weights of new layers. Gradually unfreeze selected parts of the pre-trained model, allowing comprehensive adaptation to unique regional patterns while retaining beneficial pre-trained knowledge.

The California and Puget Sound regions have input features that are closely aligned with the NHTS design in terms of the number of attributes, coding schemes, and activity types. However, their HTS datasets are much smaller than NHTS, with 53{,}823 and 7{,}791 training activity chains respectively, as in Table~\ref{tab:train_settings}. For these regions, we therefore reuse the NHTS encoder and decoder and apply only Steps 2 and 3 of the transfer procedure to fine tune the model on the regional samples.

In contrast, the Mexico City survey provides a relatively large number of activity chains (185{,}233 training chains; see Table~\ref{tab:train_settings}) but only 40 of the 60 input attributes used in NHTS and a different activity type with 10 locally defined activity categories. For Mexico City we apply all three steps of the transfer procedure. We keep the shared encoder and temporal components from the NHTS model, fine tune the linear layer after embedding to match the reduced attribute set, and replace the final activity classification layer with a new head that follows the local 10 category taxonomy before fine tuning on the regional data.

\subsection{Activity Location Assignment and Network Traffic Loading}
To evaluate the Deep Activity model in a real-world transportation network and test its applicability for transportation system analysis, we propose an activity location assignment (ALA) method as an extension of the mobility pattern generation to enhance the model’s functionality and ensure its practical application. This method aims to address the common limitation in travel survey data, where precise location information is often missing. The goal is to develop a simplified location assignment method for rapid implementation in regions without high-resolution location data.

\begin{algorithm}
\small
\caption{Activity Location Assignment}
\footnotesize
\begin{algorithmic}
\Input (1) Fitted distributions for all sub-regions: home-work/school distance distribution $\boldsymbol{D_{md}} = \{D^{1}_{md}, D^{2}_{md}, ..., D^{n}_{md}\}$; Non-mandatory trip distance $\boldsymbol{D_{nmd}} = \{D^{1}_{nmd}, D^{2}_{nmd}, ..., D^{n}_{nmd}\}$; location angular difference distribution $\boldsymbol{D_{ad}} = \{D^{1}_{ad}, D^{2}_{ad}, ..., D^{n}_{ad}\}$; (2) Land use types of all zones $\boldsymbol{LU}$; (3) Zone-to-zone distance matrix $\boldsymbol{M_{d}}$, angle matrix $\boldsymbol{M_{a}}$; (4) Home locations $\boldsymbol{HL}$ and predicted activity chains $C$ for all agents.
\Output Assigned zone IDs $\boldsymbol{Z}=\{Z^{md},Z^{nmd}\}$ for all activities, including mandatory and non-mandatory activities.
\begin{algorithmic}[1] 
\Repeat
    \For{each agent i with mandatory activities}
        \State $Dist_i = sample(\boldsymbol{D_{md}})$ \Comment{Assign home-work/school distance based on agent $i$'s zonal distance distribution}
        \State $Z^{md}_i = match(HL_i,Dist_i,LU)$ \Comment{Select the most matched zone based on assigned distance from work/school zones and land use type}
    \EndFor
    \For{each non-mandatory trip j}
        \State $Dist_j = sample(\boldsymbol{D_{nmd}})$ \Comment{Assign distance to next zone based on last activity's zonal $D_{nmd}$}
        \State $Ang2Anc_j = GetAngle(\boldsymbol{M_{a}})$ \Comment{Get the angle between previous zone and next anchor zone from angle matrix}
        \State $AngD_j = sample(\boldsymbol{D_{ad}})$ \Comment{Assign angle difference based on last activity's zonal $D_{ad}$}
        \State $Ang_j = Ang2Anc_j + AngD_j $ \Comment{Compute the direction to next location}
        \State $Z^{nmd}_j = match(Dist_j,Ang_j,LU)$ \Comment{Select the most matched zone based on the assigned distance, angle to next zone, and land use type}
    \EndFor
    \State Adjust the parameters of $\boldsymbol{D_{md}}$, $\boldsymbol{D_{nmd}}$, and $\boldsymbol{D_{ad}}$ by a small margin to slightly change the shape of the distributions
\Until{$|N_{target} - N_{assigned}| < threshold$} \Comment{Stop when the error of activity numbers across sub-regions between assigned and target locations is lower than threshold}
\State \Return $\boldsymbol{Z}$
\end{algorithmic}
\label{Algorithm1}
\end{algorithmic}
\end{algorithm}

\begin{figure}[h]
  \centering
  \includegraphics[width=0.49\textwidth]{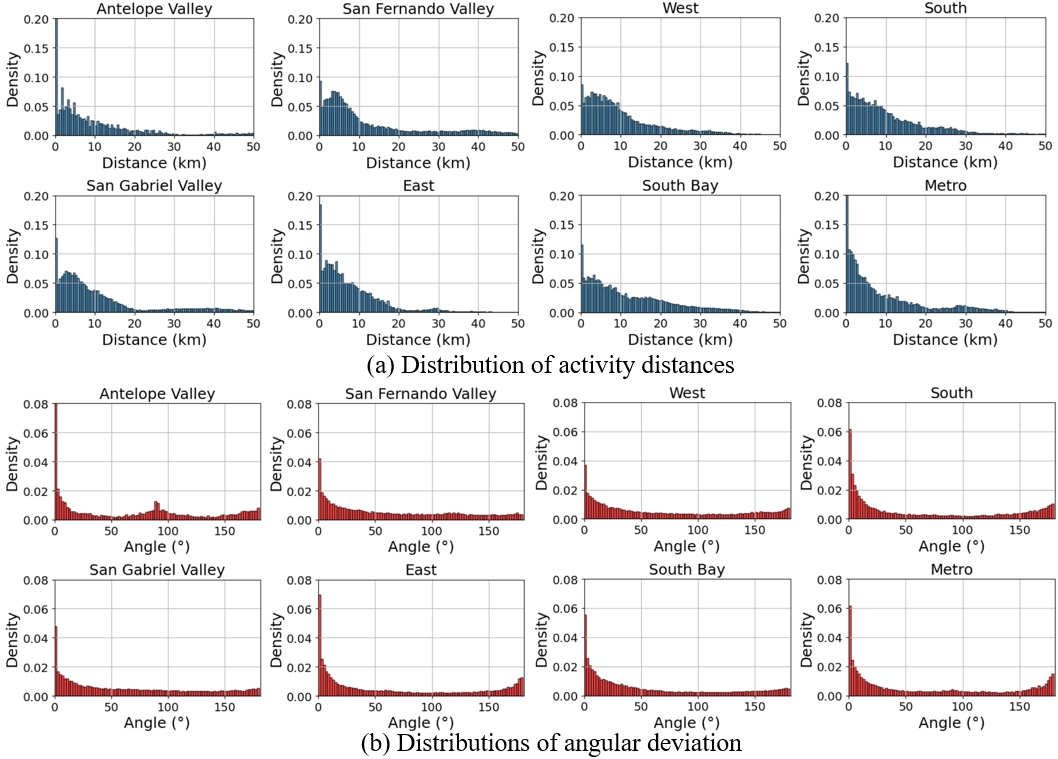}
  \caption{Distributions of activity distances and angular deviations across sub-regions in LA}
  \label{fig:distribution}
  \vspace{-2mm}

\end{figure}

The proposed method assigns zone-level locations $Z$ for each predicted activity by considering the distribution of distances and angular deviations between preceding and subsequent activities. For large metropolitan regions, the spatial distribution is further refined by applying sub-region-specific distance and angle distributions to capture local spatial variations. This ensures that the assigned locations reflect the heterogeneity within different areas of the region. This is particularly important in large metropolitan areas, such as the Greater LA area, where spatial distributions may vary significantly across sub-regions, as shown in Fig. \ref{fig:distribution}.

The proposed ALA process consists of the following key steps, as shown in \textbf{Algorithm 2}:

\textbf{Step 1. Assigning TAZs for Mandatory Activities.}
The primary assumption is that each individual’s home location is predetermined in the dataset. The first group of activities to be assigned locations are mandatory activities (e.g., work or school). For each individual, a home-to-work or home-to-school distance is assigned based on their demographic characteristics. Within each sub-region, these assigned distances follow the target distribution of commute distances between home and mandatory activity locations for that specific sub-region, denoted as $\boldsymbol{D_{md}}$. A TAZ with the appropriate land use type (work or school) and closely matching the assigned commute distance is then allocated to the individual.

\textbf{Step 2. Assigning TAZs for Non-Mandatory Activities.}
After assigning TAZs for mandatory activities, these locations serve as anchor points within the individual’s daily activity chain. The subsequent step involves assigning locations for non-mandatory activities (e.g., shopping, leisure, exercise) that occur between these anchor points. Non-mandatory activity locations are assigned based on two key parameters: (1) the distance to the next non-mandatory activity, and (2) the angular deviation between the direct path to the next non-mandatory location and the direct path to the next anchor location. The assigned distances and angular deviations follow the target distributions of distance ($\boldsymbol{D_{nmd}}$) and angular deviations ($\boldsymbol{D_{ad}}$) for each sub-region, ensuring that the spatial distribution aligns with regional patterns.

\textbf{Step 3. Refinement of Location Assignment.}
The objective of the location assignment process is to ensure that the spatial distribution of generated activity locations closely resembles the true activity distribution across sub-regions. This spatial similarity is assessed by comparing the occurrence frequencies of activities across sub-regions in the location assignment output with those in the ground truth data. To minimize bias in the generated distribution, the reference distributions used in Steps 1 and 2 are iteratively adjusted until the assigned activity number of sub-regions match the ground truth. This refinement process is initially applied to a small sample of the population to fine-tune the reference distributions for each sub-region. The refined distributions ($\boldsymbol{D_{md}}$, $\boldsymbol{D_{nmd}}$, and $\boldsymbol{D_{ad}}$)  are then applied to the larger population for large-scale transportation system analysis.

It is important to note that the objective of this location assignment method is not to predict the exact location of each activity, but rather to ensure a realistic spatial distribution of activity locations across the study area. This method provides a suitable input for subsequent transportation system analyses, such as regional traffic volume estimation and congestion assessment.

The mobility patterns and the corresponding activity location assignments together form a comprehensive regional travel demand input. This generated travel demand is integrated into an existing transportation simulation framework, LASim~\cite{he2024multilasim}. LASim is a large-scale, agent-based multimodal transportation simulation designed for the Greater LA area, as shown in Fig. \ref{fig:la_map} (c). The framework builds on the Multi-Agent Transport Simulation (MATSim) to tackle the challenges induced by urbanization and changing mobility patterns. By loading the synthesized travel demand into the LA roadway network, we can generate the synthetic traffic flow and further evaluate the transportation system performance based on the generated human mobility patterns.

\section{Experiments and Results}
\subsection{Training and Evaluation Methods}
All experiments are conducted on an NVIDIA RTX A5000 GPU. The datasets, their train, validation, and test splits, and the common training configuration used for the NHTS base model and for the three regional fine-tuning experiments are summarized in Table~\ref{tab:train_settings}. As described in Section~V-D, we first train a generic base model on the nationwide NHTS data and then adapt it to three regional datasets from California, the Puget Sound region, and Mexico City using transfer learning.

\begin{table}
\caption{Datasets and training configuration for the base model and regional fine tuning}
\label{tab:train_settings}
\centering
\begin{tabular}{l|cccc}
\hline
 & NHTS (US) & California & Puget Sound & Mexico City \\
\hline
Train Data
& 160{,}831 
& 53{,}823 
& 7{,}791 
& 185{,}233 \\

Val. Data 
& 18{,}106 
& 7{,}889 
& 1{,}113 
& 23{,}154 \\

Test Data
& 18{,}106 
& 15{,}378 
& 2{,}226 
& 23{,}154 \\
\hline
Optimizer 
& \multicolumn{4}{c}{Adam} \\

Initial LR 
& \multicolumn{4}{c}{0.005} \\

LR decay 
& \multicolumn{4}{c}{$\times 0.95$ per epoch} \\

Batch size 
& \multicolumn{4}{c}{1024} \\

Epochs 
& \multicolumn{4}{c}{150 or early stop if convergency is observed} \\
\hline
\end{tabular}
\end{table}

The inherent uncertainty and variability in human behavior make it challenging and often inappropriate to evaluate the accuracy of a specific agent. Moreover, data synthesis differs fundamentally from prediction: the goal is not to replicate exact individual trajectories from the NHTS dataset. Doing so would merely reproduce the original data, undermining the purpose of generating novel synthetic agents. Therefore, the Deep Activity Model is trained on large batches of agents to help it learn the underlying behavioral distributions conditioned on socio-demographic inputs. During inference, it generates diverse yet behaviorally consistent activity chains, enabling realistic agent-level mobility without replicating observed data. Accordingly, we evaluate the model based on population-level distributional fidelity—such as activity frequencies, temporal patterns, and transition probabilities—rather than single-agent classification accuracy. This approach aligns with the goal of creating synthetic populations for downstream simulation and planning applications.

In this paper, the Jensen-Shannon Divergence (JSD) is used as the similarity metric~\cite{luca2021survey}, as shown in Equation \ref{eqJSD}. The goal is to minimize the difference between the distributions of generated and real activity patterns from activity chains. The metrics include: 1) \textbf{activity frequencies}, 2) \textbf{start times}, 3) \textbf{end times}, 4) \textbf{number of daily activities} (activity chain length), and 5) \textbf{duration of each activity}.

\begin{align}\label{eqJSD}
JSD(P \| Q)=\frac{1}{2} \sum_{x \in X}\left[P(x) \log \left(\frac{P(x)}{M(x)}\right)\right] \notag \\
+\frac{1}{2} \sum_{x \in X}\left[Q(x) \log \left(\frac{Q(x)}{M(x)}\right)\right]
\end{align}

\noindent where $M=(P+Q)/2$. Here, $P$ is the distribution of activity patterns from the generated activity chains, and $Q$ is the distribution from the ground truth activity chains. $X$ represents the full range of probabilities for a specific activity pattern statistic. A JSD value closer to zero indicates greater similarity between the distributions, showing the model's effectiveness in approximating the true distribution.

In addition to quantifying the model's performance at the distribution level using JSD values, it is also critical to analyze it at the chain level. By aggregating activity chains into graphs, where activity types are nodes and transitions between activities are edges, \textbf{completeness} of the activity chains, as the sixth metric, can be examined to ensure no activity types or transitions are missing. These graphs can be converted into transition matrices, which reveal the transition probabilities between any two pairs of activities. As the seventh metric, the \textbf{similarity of activity transition probability} is quantified by \textbf{Frobenius norm}, i.e., $|A-B|_F=\sqrt{\sum_{l=1}^L \sum_{m=1}^M\left|a_{l m}-b_{l m}\right|^2}$, where matrix $A$ and $B$ are the transition matrix of generated activity chains and the ground truth, respectively.

Besides activity-generation evaluation, we validate the ALA and the subsequent network traffic loading at the transportation-system level. Because the goal of the location assignment module is to reproduce realistic spatial demand patterns rather than to predict the exact next zone in each activity chain, we do not use standard location-prediction metrics such as top-$k$ recommendation accuracy~\cite{liu2016predictingtopk}. Instead, we benchmark the synthetic demand against the SCAG ABM using system-level indicators: the cosine similarity of origin–destination (OD) flow matrices~\cite{guo2020odcosine1}, the spatial distribution of activities across sub-regions~\cite{mungthanya2019constructingcosine2}, and hourly vehicle-miles-traveled (VMT)~\cite{jiang2022connected}. These comparisons quantify how closely the generated demand and assigned locations reproduce the behavior of the existing regional planning model. To assess external realism beyond this calibrated baseline, we further compare simulated traffic conditions with real-world observations from Caltrans PeMS on the I-405 corridor, focusing on whether the temporal congestion patterns are consistent across our simulation, the SCAG ABM, and PeMS sensors. At the city-network level we report Mean Absolute Percentage Errors (MAPE) between our model and SCAG ABM only, treating the ABM as the calibrated regional demand model. At the corridor level we again compute MAPE for traffic volume and speed relative to SCAG ABM, while using PeMS curves as an observational ground truth reference for congestion timing and severity rather than as the direct target for error minimization.

In summary, our evaluation strategy follows the three stages of the pipeline in Fig.~\ref{fig:system_workflow}. First, for activity generation, we compare population-level distributions of generated activity chains against HTS data test set, using the NHTS for the generic model and regional HTS for the transfer-learning experiments. Second, for spatial demand and system-level metrics in LA County, we benchmark the assigned activity locations and hourly VMT against the SCAG ABM, which provides zonal locations and a calibrated regional travel demand pattern. Third, for traffic realism, we load the generated demand into the LASim agent-based simulation and compare the resulting traffic volumes and speeds with Caltrans PeMS observations on the I-405 corridor, using those obtained when loading SCAG ABM demand as an external reference for congestion timing and severity rather than as a direct calibration target.

\begin{figure*}[ht]
  \centering
  \includegraphics[width=0.95\textwidth]{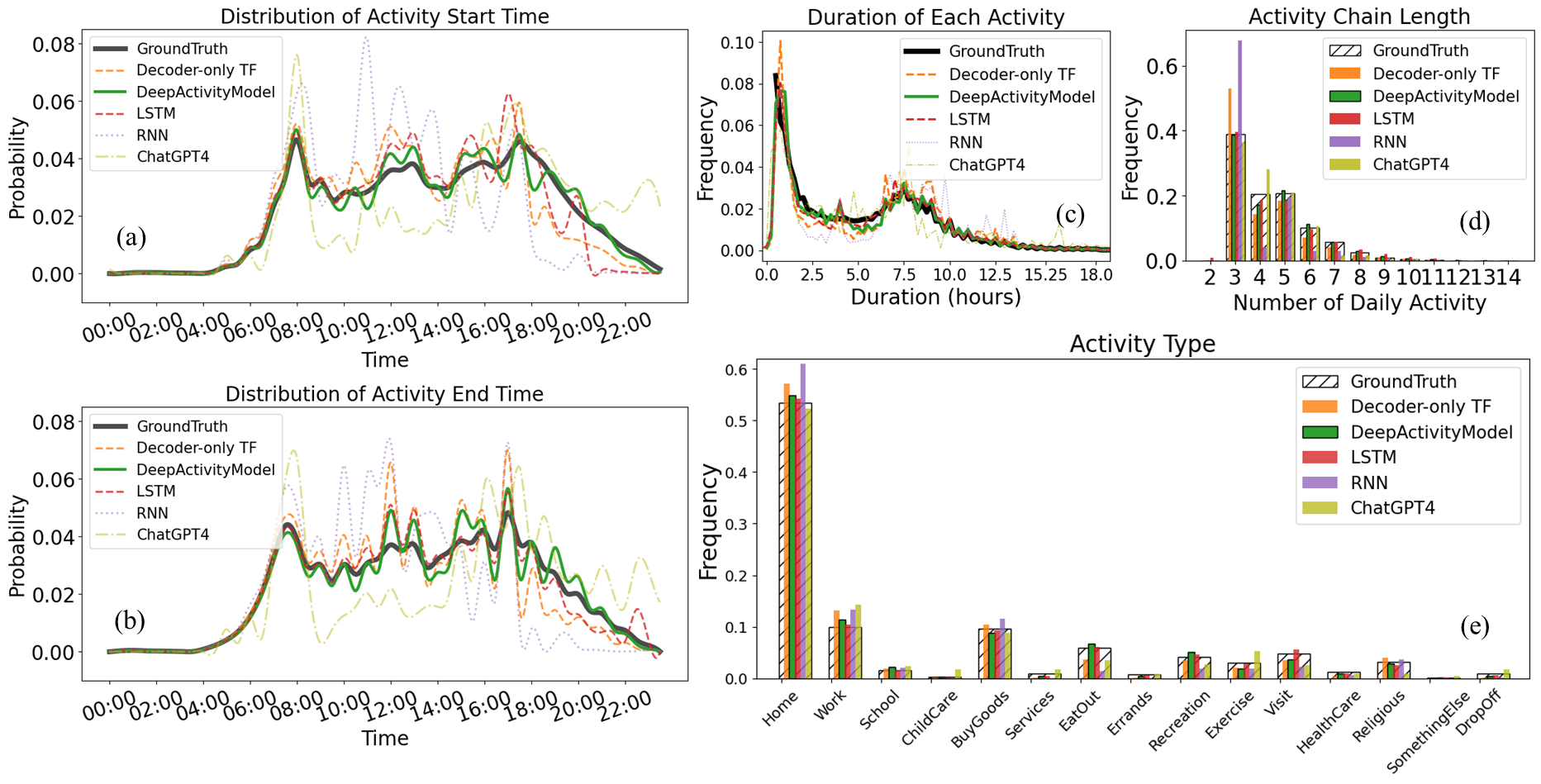}
  \caption{Detailed analysis and comparison of activity generation on (a)(b) temporal dynamics, (c) activity chain length, (d) activity duration, and (e) activity type distribution.}
  \label{fig:resultMain}
  \vspace{-3mm}
\end{figure*}

\subsection{Baseline Models}
\textbf{Decoder-only transformer}. In language modeling and sequence generation tasks, the "decoder-only transformer" is commonly adopted due to its effectiveness~\cite{roberts2023computational}, making it a natural baseline for activity generation, allowing for a clear comparison when evaluating more complex models. 

Recurrent Neural Network (RNN) and their variants are widely used for predicting travel behavior, e.g., next location prediction \cite{al2016stf}. In this study, they are used as a comparison against the transformer-based models. 

\textbf{Vanilla RNN} has a simple architecture where the output from the previous step is fed back into the network to influence the output of the current step \cite{al2016stf}. 

\textbf{Gated Recurrent Unit (GRU)} introduces gating mechanisms to control the flow of information, maintain long-term dependencies, and address the vanishing gradient problem \cite{liao2018predicting}.

\textbf{Long Short-Term Memory (LSTM)} features a more complex architecture, consisting of three gates: the input gate, the forget gate, and the output gate \cite{krishna2018lstm}, compared to the two gates used by GRU.


\textbf{Large Language Models (LLMs)} serve as suitable baselines for human mobility modeling due to their exceptional capabilities in understanding context and generating complex sequences without extensive training periods. In our previous study \cite{Liu2024HumanMobility}, we utilized pre-trained models such as OpenAI's GPT-4 \cite{hurst2024gpt}, Meta's LLaMA3.1-70B \cite{dubey2024llama}, and DeepSeek-R1 \cite{guo2025deepseekr1} to generate daily activity chains based solely on socio-demographic information, without the need for long-term training on domain-specific data.

All LLM-based baselines were prompted using a structured design incorporating the task description, high-level statistical context, generation instructions, few-shot examples, and retrieval-based augmentation, following the methodology established in prior work \cite{Liu2024HumanMobility}. Decoding parameters adhered to each model provider's API defaults. Temperature was set to 1.0 and top-p to 1.0, allowing full token distribution sampling. Frequency and presence penalties remained at their defaults of 0.0. No explicit repetition penalty, top-k sampling, or post-generation constraints were applied. Truncation behavior, maximum token limits, and the number of outputs (n = 1) followed provider defaults. Typical prompt input lengths were approximately 700 tokens, with generated activity chains averaging around 4 tokens each. The generation speed was approximately 0.5 seconds per instance for GPT-4 and DeepSeek-V3 API calls, and 1 second for LLaMA3.1-70B with 4-bit quantization running on an L40S GPU with 48GB memory. At these rates, the estimated cost is approximately US\$0.8 for every 1,000 activity chains generated.

\subsection{Evaluation on Activity Generation}
\subsubsection{Distribution Similarity}
To evaluate the performance of the proposed Deep Activity model, a comparative analysis is conducted involving the baseline models. The results of seven metrics are presented in Table \ref{table:comparison}, where the proposed Deep Activity model outperforms the others by achieving the lowest JSD values for activity chain length, duration, start time, and end time. These results indicate a high degree of similarity between the generated and ground truth activity patterns, underscoring the model's accuracy in capturing dynamic human activities. Additionally, it excels in edge completeness, with a percentage of 92.2\%, significantly surpassing other models and illustrating its robustness in capturing the full spectrum of activity transitions. The LSTM model stands out in two metrics, showing superior performance compared to the Decoder-only Transformer. It achieves the lowest JSD value for activity type and the lowest Frobenius norm, indicating minimal discrepancy in transition probabilities between the generated and ground truth activity chains. 

Although the overall temporal fidelity is high (JSD = 0.003), Fig.\ref{fig:resultMain} (a) and (b) reveal a localized overestimation of activity endings between 10:00 and 15:00, with a higher JSD of approximately 0.005. This discrepancy, while still within acceptable bounds for practical modeling, is primarily driven by non-mandatory activities (e.g., shopping, dining, errands), which show greater variability than mandatory ones. The model more easily captures the regularity of work and school, whereas the flexible timing of midday activities introduces added uncertainty.


The LLMs demonstrate mixed results. While LLMs like GPT-4 and DeepSeek-R1 show promise in activity type and chain length prediction, their higher JSD values for temporal aspects and lower edge completeness scores reveal significant limitations in capturing the complex dynamics of daily activities. DeepSeek shows competitive performance in temporal aspects (start and end times) with relatively low JSD values of 0.013, but struggles with activity chain length (JSD = 0.040) and has limited edge completeness (43.2\%). These constraints, particularly evident in LLaMA3's underperformance across all metrics, indicate that current LLMs are not yet suitable for generating accurate activity chains without substantial adaptations to better model the nuanced patterns of human routines.

On the other hand, traditional models like GRU and RNN, with notably higher JSD values, demonstrate moderate performance but fall behind more advanced approaches. This indicates that they struggle to capture the complex temporal dependencies and transitions that characterize human activity patterns.

\begin{table}[ht]
\centering
\caption{Activity chain generation evaluation}
\footnotesize
\begin{tabular}{@{}lccccccc@{}}
\toprule
\textbf{Model} & \textbf{Len.} & \textbf{Dur.} & \textbf{Start} & \textbf{End} & \textbf{Type} & \textbf{EC} & \textbf{F-Norm} \\
\midrule
GRU & 0.015 & 0.032 & 0.085 & 0.217 & 0.013 & 51.6\% & 0.934 \\
RNN & 0.064 & 0.029 & 0.067 & 0.121 & 0.024 & 50.9\% & 1.291 \\
LSTM & 0.006 & 0.003 & 0.019 & 0.004 & \textbf{0.003} & 91.4\% & \textbf{0.377} \\
D-TF & 0.011 & 0.011 & 0.014 & 0.012 & 0.013 & 67.1\% & 0.784 \\
GPT-4  & 0.011 & 0.018 & 0.064 & 0.074 & 0.009 & 42.4\% & 1.111 \\
LLaMA3 & 0.011 & 0.023 & 0.090 & 0.081 & 0.024 & 45.9\% & 1.203 \\
DeepSeek & 0.040 & 0.013 & 0.013 & 0.026 & 0.009 & 43.2\% & 0.987 \\
Proposed & \textbf{0.002} & \textbf{0.002} & \textbf{0.003} & \textbf{0.003} & 0.005 & \textbf{92.2\%} & 0.643 \\
\bottomrule
\end{tabular}
\footnotesize *D-TF: decoder-only transformer; Len.: activity chain length; Dur.: duration of each activity; EC: edge completeness in percentage; F-Norm: Frobenius norm. All models reach 100\% node (activity occurrence) completeness. Numbers except EC and F-Norm are JSD values.
\label{table:comparison}
\end{table}

In addition to the quantitative analysis, Fig. \ref{fig:resultMain} provides a deeper insight into how each model performs in predicting specific details of human activities. Poorly performing models were excluded from the figure to maintain focus on the most relevant comparisons. For instance, in terms of start times (in Fig. \ref{fig:resultMain}(a)), both the LSTM and Decoder-only Transformer underestimate evening activities, whereas for end times (in Fig. \ref{fig:resultMain}(b)), the LSTM overestimates evening activities, and the Decoder-only Transformer overestimates midday activities. Meanwhile, ChatGPT4 demonstrates high volatility in predicting both start and end times, suggesting challenges in capturing consistent temporal patterns. Regarding the activity chain length (in Fig. \ref{fig:resultMain}(c)), the Decoder-only Transformer tends to generate three activities per day for individuals, while all models, except ChatGPT4, tend to underestimate the four-activity chains. This suggests a tendency in most models to simplify daily activity sequences, potentially missing the complexity of real-world behavior. In terms of activity types (in Fig. \ref{fig:resultMain}(e)), Transformer-based models, LSTM, and ChatGPT4 perform well, especially for the common activities.

\subsubsection{Loss Term Ablation Study}
To assess the impact of individual loss terms in our Deep Activity model, we performed an ablation study, summarized in Table \ref{tab:lossAblation}. The findings illustrate how each loss term contributes to model performance across the seven metrics.

Removing the \textbf{soft label loss ($L_s$)} has the most significant impact across all metrics. This underscores the critical role of $L_s$ in encouraging the model to explore different combinations and learn overall temporal patterns, rather than overfitting to specific time stamps. The flexibility provided by $L_s$ appears crucial for capturing the inherent variability in human activity schedules. The absence of the \textbf{temporal order loss ($L_o$)} results in noticeable performance drops across all metrics, including duration and activity chain length. While the impact on start and end time predictions is less severe than removing $L_s$, the decline in length accuracy suggests that $L_o$ plays a role in maintaining not just the chronological order, but also the overall structure of daily activity chains. When the \textbf{sequential timing loss ($L_{seq}$)} is removed, we observe relatively minor decreases in most metrics, with duration accuracy remaining largely unchanged. However, the higher Frobenius norm indicates that $L_{seq}$ is particularly important for maintaining accurate activity transition probabilities, even if its impact on individual activity timings is less pronounced.

\begin{table}[h]
\centering
\caption{Influence of each loss term}
\begin{tabular}{@{}lccccccccc@{}}
\toprule
\textbf{Loss} & \textbf{Len.} & \textbf{Dur.} & \textbf{Start} & \textbf{End} & \textbf{Type}  & \textbf{EC} & \textbf{F-Norm} \\ \midrule
w/o $L_s$         & .025  & .012    & .039      & .024    & .018 & 65.6\%       & .907                \\
w/o $L_o$     & .024  & .011    & .017      & .016    & .014 & 68.5\%       & .711                \\
w/o $L_{seq}$  & .013  & .006    & .015      & .014    & .013 & 72.6\%      & .739                \\
All                   & .002  & .002    & .003      & .003    & .005 & 92.2\%       & .643                \\ \bottomrule
\end{tabular}
\label{tab:lossAblation}
\vspace{-2mm}

\end{table}

\subsubsection{Contextual Variation}
\begin{figure}[ht]
  \centering
  \includegraphics[width=0.49\textwidth]{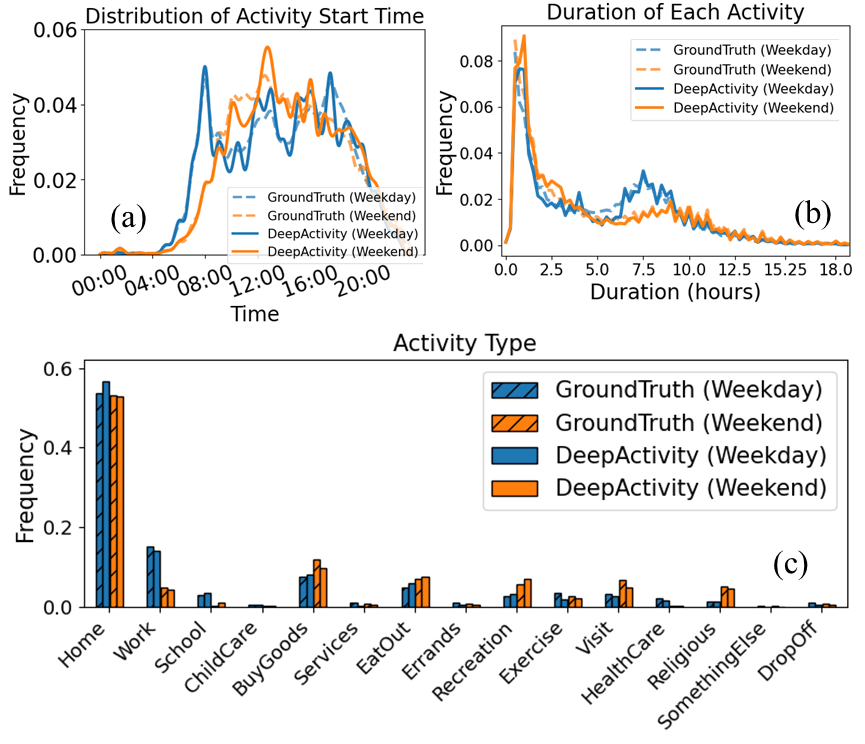}
  \caption{Activity patterns in weekdays and weekends}
  \label{fig:WeekendWeekday}
\end{figure}

Distinct activity patterns between weekdays and weekends are effectively captured by the proposed model, as presented in Fig. \ref{fig:WeekendWeekday}. The start time distribution (a) reveals a later peak for weekend activities compared to weekdays, with a notable weekend shift towards midday starts. In Fig. \ref{fig:WeekendWeekday}(b), there are more activities around 7.5 hours during weekdays, implying the working hours, which can also reflected in Fig. \ref{fig:WeekendWeekday}(c). Clear variations are displayed in activity types, with work-related activities dominating weekdays while leisure activities like "EatOut" and "Visit" increase on weekends. Notably, the proposed model closely mirrors these temporal and categorical differences, demonstrating its ability to distinguish and reproduce weekday-weekend variations in human activity patterns.

\begin{figure}[ht]
  \centering
  \includegraphics[width=0.49\textwidth]{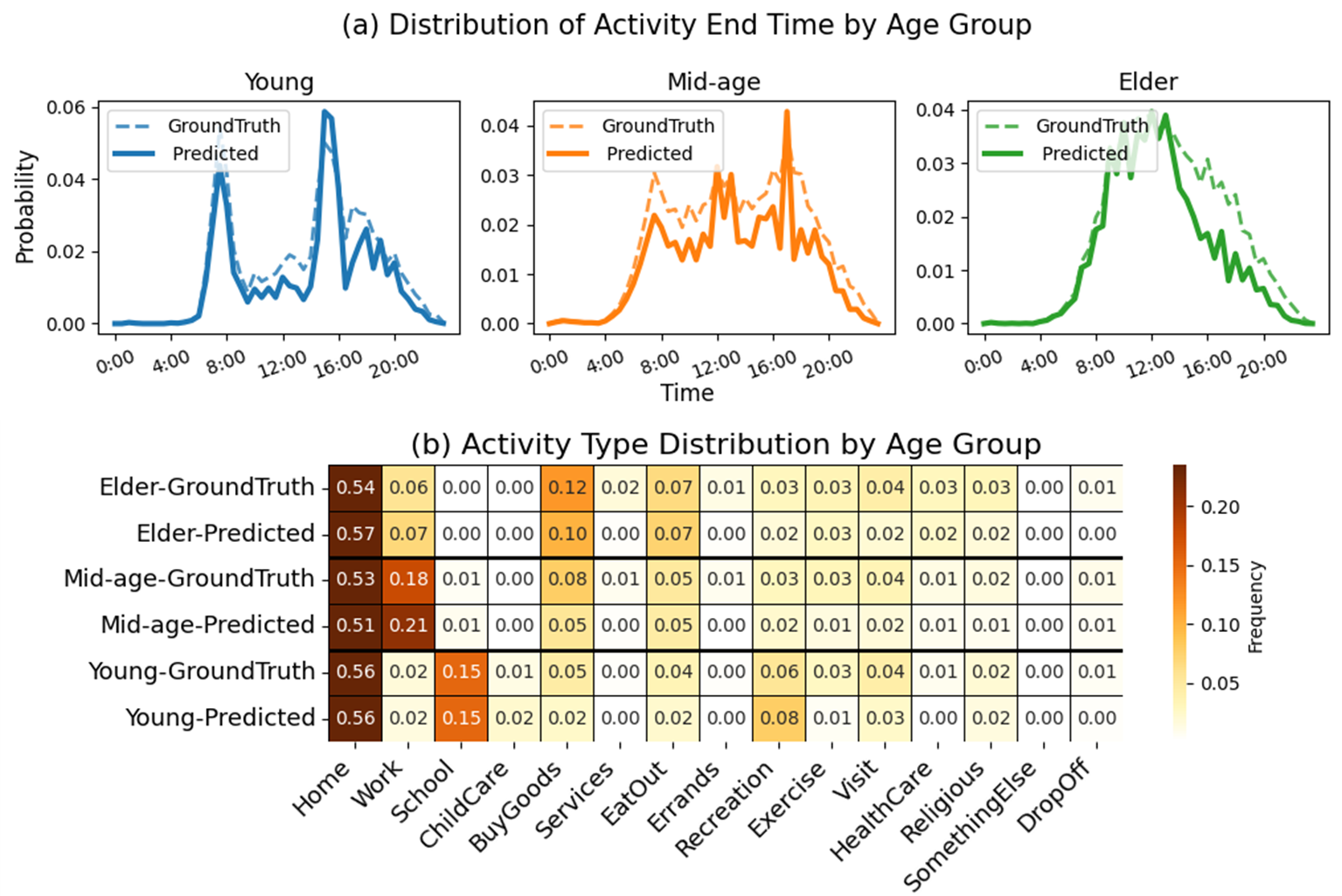}
  \caption{Activity patterns across age groups}
  \label{fig:AgeResult}
  \vspace{-3mm}

\end{figure}

Age is another crucial factor influencing activity patterns synthesis process, alongside day-of-week. Fig. \ref{fig:AgeResult} illustrates how the proposed model captures age-related differences across young (0-18), middle-aged (19-65), and elderly (65+) groups. In Fig. \ref{fig:AgeResult} (a), distinct end-time distributions are observed. Young individuals show peaks aligned with school schedules, middle-aged adults exhibit a more varied pattern reflecting diverse work commitments, while the elderly display a gradual curve peaking around midday. The activity type distribution in Fig. \ref{fig:AgeResult} (b) further highlights these differences, with high school attendance and recreation for the young, significant work-related activities for middle-aged, and increased shopping and eating time for the elderly. Notably, the Deep Activity model accurately reproduces these age-specific patterns in both timing and activity types, demonstrating its ability to synthesize realistic activity chains that reflect the distinct lifestyles associated with different age groups.

\subsubsection{Interdependency among Household members and Activity in Activity Generation}\label{sec:explanability}
\begin{figure}
  \centering
  \subfigure[Attention weights]{
    \includegraphics[width=0.99\linewidth]{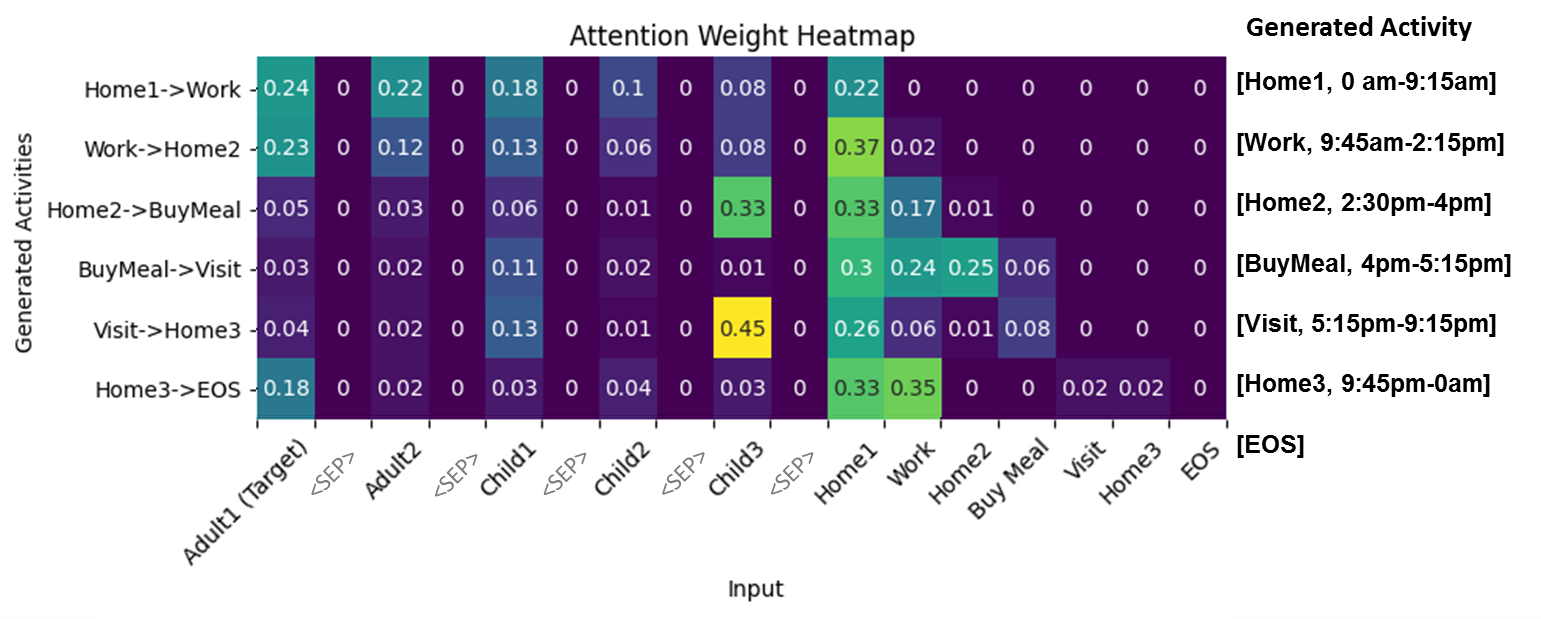}
    }
  \hfill
  \subfigure[Activity chain generation process]{
    \includegraphics[width=0.8\linewidth]{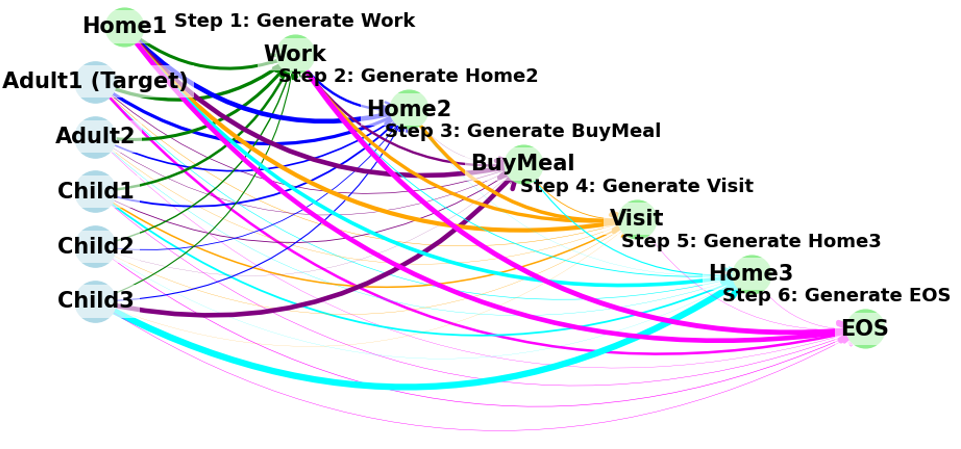}
    }
  
  \hfill
  \subfigure[Attention weights if child 3 was removed.]{
    \includegraphics[width=0.99\linewidth]{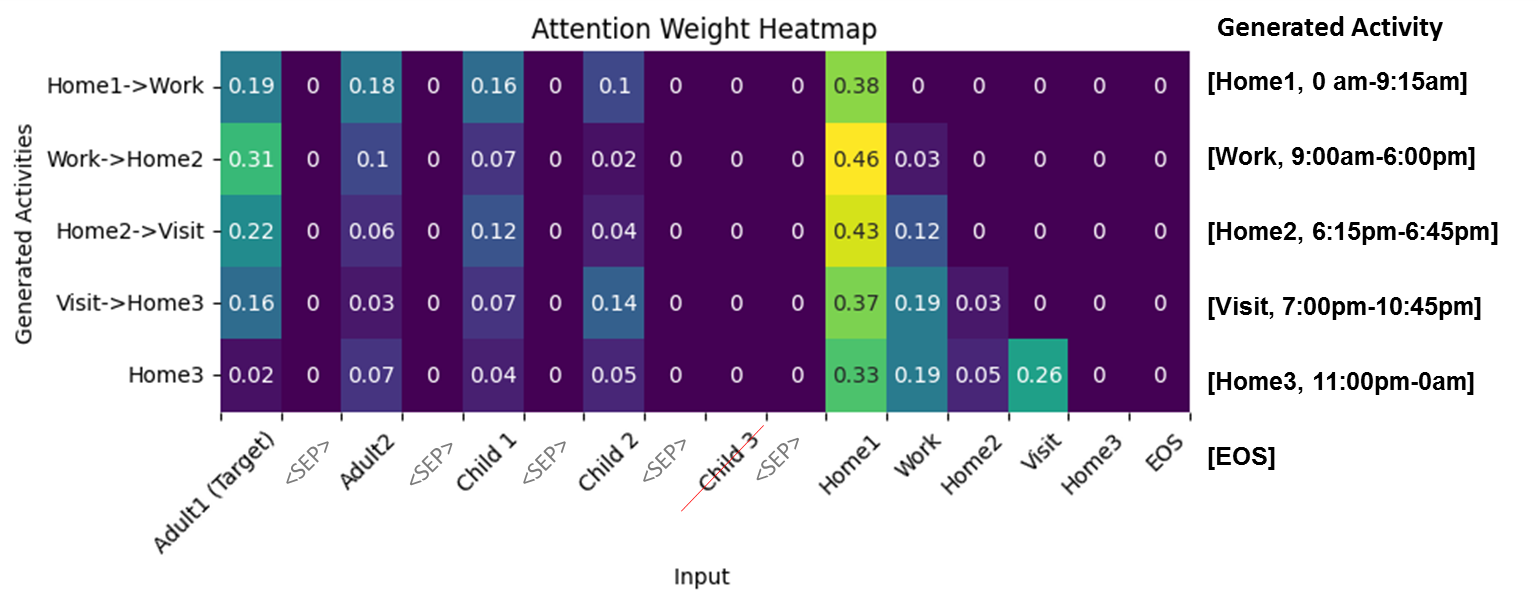}
    }
  \hfill
  
  \subfigure[Activity chain generation process if child 3 was removed.]{
    \includegraphics[width=0.85\linewidth]{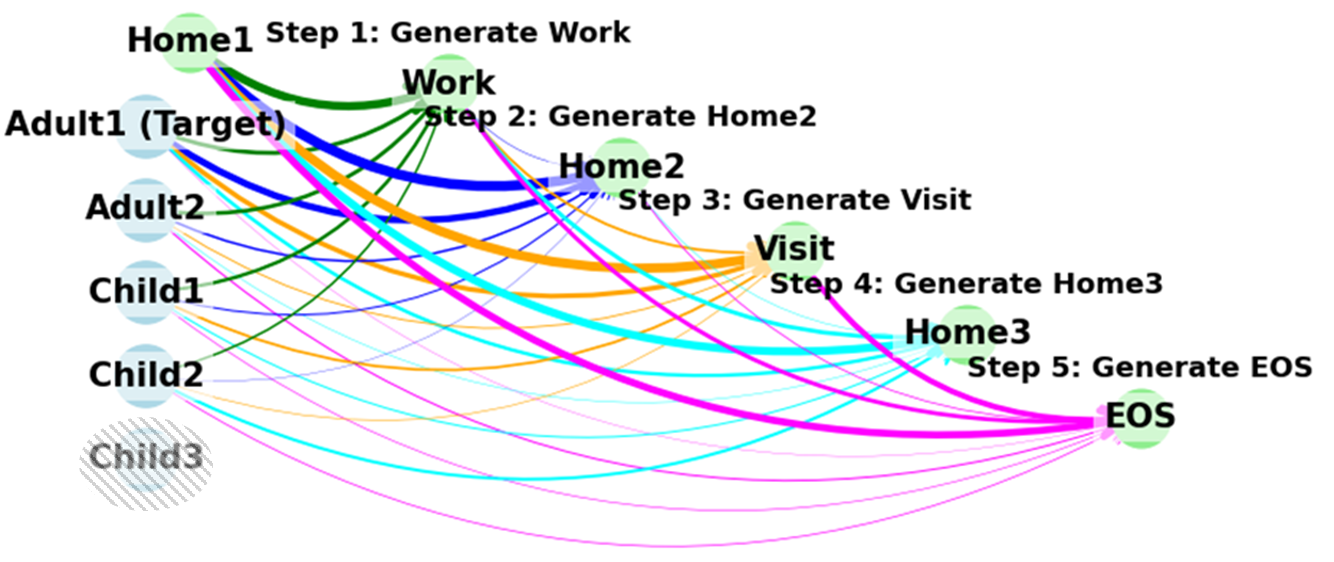}
    }
  \caption{Attention weights reveal the interdependency among household members and activities.}\label{fig:resultATT}  
  \vspace{-2mm}

\end{figure}

The utilization of attention mechanisms in transformer models provides valuable insights into the decision-making processes within families, as illustrated by the detailed attention heatmap and graphical representation in Fig. \ref{fig:resultATT}. With its unique input data design, we are able to visit the attention from the first layer of encoder of the Deep Activity model, which reveals the interdependencies among the person's household and their activities.

Each row in the heat map (Fig. \ref{fig:resultATT}(a)) corresponds to specific activities of the target person—a male worker and father in a five-member family. The columns demonstrate how interactions with other family members and previous activities influence subsequent activities. The step-by-step activity generation is detailed in Fig. \ref{fig:resultATT}(b), where each step is labeled in a unique color, and the varied thickness of the lines indicates the relative influence of each interaction. For example, Child 3 exerts a significant influence on the target person's activities, such as "BuyMeal" and "Visit," highlighting the interdependencies of family members in coordinating daily schedules. To provide direct empirical evidence of interactions among household members, a comparative case study is presented in Fig. \ref{fig:resultATT} (c)(d). It shows the results of removing the third child from the input features of the same five-member household used in the original attention analysis. The removal leads to changes in both the attention heatmap and the generated activity chain of the target individual. In the encoder, the attention distribution shifts noticeably, with reduced weights on features related to child care and joint scheduling. The activity chain of the target person also changes in both type and timing. Compared to the full-household case shown in \ref{fig:resultATT}(a)(b), the "BuyMeal" activity is omitted, and other activities are shifted in time when the third child is excluded.

\subsubsection{Model Transferability} \label{Sec:transfer}
\begin{figure}
  \centering
  \includegraphics[width=0.49\textwidth]{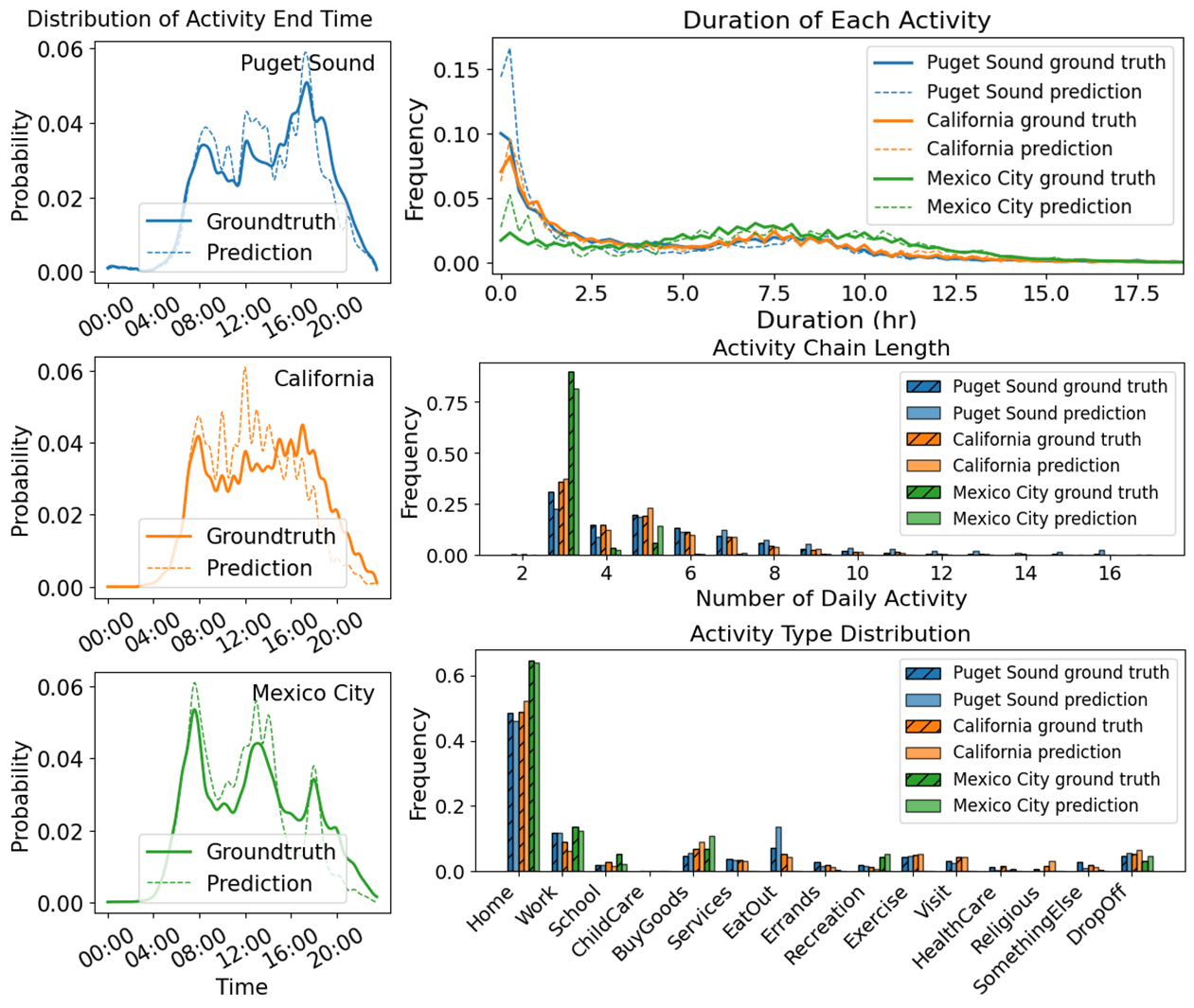}
  \caption{Distribution comparison for datasets from three regions, showing significantly different activity patterns. Activity type labels are excluded because the dataset of Mexico City only contains 10 types of activity that are different from CA and Puget Sound region.}
  \label{fig:DataComparison}
\end{figure}

To demonstrate the transferability of the proposed model, fine-tuning techniques are applied to California, Puget Sound, and Mexico City. These regions cover diverse geographies and sizes, showing notable differences in mobility patterns (Fig. \ref{fig:DataComparison}). Each dataset reveals distinct activity patterns in terms of timing, duration, type, and daily frequency. 

For activity end times, Puget Sound and California peak around 17:00-18:00, while Mexico City shows three peaks, notably at 08:00 and 14:00-15:00. Puget Sound has more short activities (less than 0.5 hours) compared to the other regions. Chain lengths vary, with Mexico City having more 3-activity chains than other regions. "Home" is the most frequent activity type across all regions, followed by "Work." Mexico City has a higher proportion of "Home" and fewer "Work" activities than the U.S. regions. Activity types like "EatOut," "ChildCare," and "Visit" present in the U.S. datasets are missing from Mexico City HTS, where "Exercise" is grouped under "Recreation." These differences highlight challenges in standardizing activity classifications, emphasizing the need for region-specific fine-tuning of the NHTS pre-trained model to capture unique mobility behaviors.

The fine-tuning approach has proven highly effective, as indicated in Table \ref{table:region_comparison}. The fine-tuned models have demonstrated the capability to transfer the generic model to different regions while maintaining robustness and achieving accuracy levels comparable to those obtained using the full NHTS dataset. 

\begin{table}[ht]
\centering
  \vspace{-1mm}
\caption{Transfered Deep Activity model to California, Puget Sound, and Mexico City}
\footnotesize
\begin{tabular}{@{}lccccccc@{}}
\hline
\textbf{Region} & \textbf{Len.} & \textbf{Start} & \textbf{End} & \textbf{Dur.} & \textbf{Type} & \textbf{EC} & \textbf{F-Norm} \\ \hline 
California & .004&.013&.033&.002&.007&83.6\%&.460\\ \hline
Puget Sound & .030&.009&.051&.010&.012&79.5\%&.652 \\ \hline
Mexico City & .010&.056&.010&.006&.009&63\%&.339\\ \hline
\end{tabular}
\label{table:region_comparison}
\vspace{-1mm}
\end{table}

To further illustrate the performance of our proposed method, we visualized detailed fine-tuning results for each region, as illustrated in Fig. \ref{fig:DataComparison}. As summarized in Table~\ref{tab:train_settings}, the fine-tuning datasets for California and Puget Sound contain 53{,}823 and 7{,}791 training activity chains, which are much smaller than the 160{,}831 chains used for NHTS pre training, whereas the Mexico City dataset provides 185{,}233 training chains but with a different set of input attributes and a different activity type. Even under these heterogeneous conditions, the fine-tuned models for all three regions still produce distributions that closely match the regional ground truth across all metrics, with only minor discrepancies. This demonstrates the model's effectiveness in capturing and reproducing diverse regional mobility patterns, validating the success of the transfer learning approach in adapting to different urban environments.

\subsubsection{Data Balancing Algorithm Evaluation}
\begin{figure}
  \centering
  \includegraphics[width=0.49\textwidth]{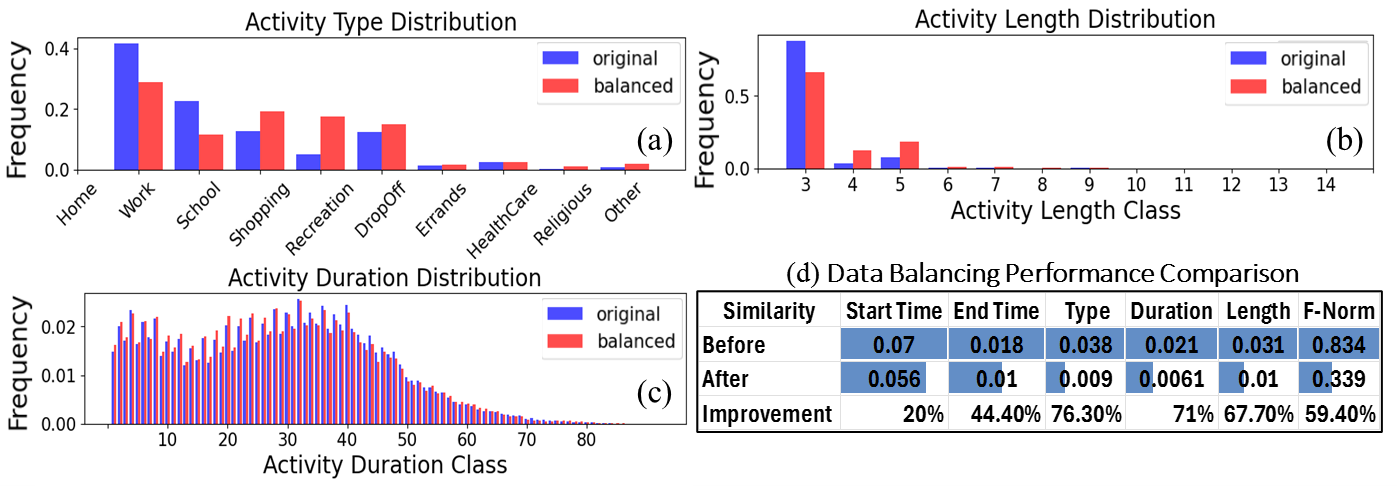}
  \caption{Data balancing performed on Mexico City.}
  \label{fig:Databalancing}
  \vspace{-2mm}

\end{figure}

: Data balancing is performed on biased datasets to mitigate issues of skewed representation. In our experiments we apply Algorithm~1 to the NHTS and Mexico City training sets, and we use the most biased one, Mexico City, to illustrate the effect of balancing in Fig.\ref{fig:Databalancing}, especially for activity length and activity type. The balancing process reduces the overrepresentation of common activities such as work and school, while enhancing the visibility of less frequent activities. It also reduced the dominance of 3-activity chains, helping to achieve a more even activity distribution. Importantly, while other features were adjusted, the overall activity duration distribution remained consistent. For better visualization, the dominant "Home" activity was excluded.

The outcomes of this balancing effort is elaborated in Fig. \ref{fig:Databalancing}. Fig \ref{fig:Databalancing}. (a) and (c) illustrate the adjusted distributions for activity type and length, with a noticeable reduction in overrepresented activities and chains. Fig\ref{fig:Databalancing}. (b) highlights that the activity duration distribution is preserved despite other changes. Fig \ref{fig:Databalancing}. (d) depicts improvements in model performance metrics, showing significant reductions in JSD for activity type, from 0.038 to 0.009, indicating over 76.3\% improvement. Frobenius Norm is also improved by 59.4\%, from 0.834 to 0.339. Though temporal metrics show smaller enhancements due to the focus on activity duration, overall model performance benefits significantly from data balancing, demonstrating its effectiveness in mitigating dataset biases.

\subsection{Validation of ALA and Network Traffic Loading}

\begin{figure}
  \centering
  \includegraphics[width=0.49\textwidth]{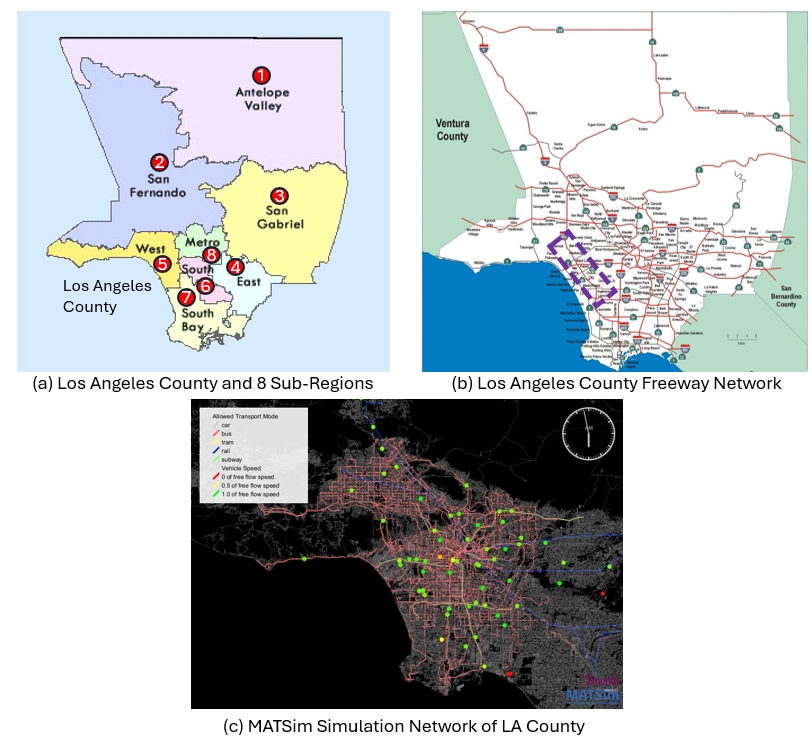}
  \caption{LA county map and freeway network.}
  \label{fig:la_map}
  \vspace{-1.5mm}

\end{figure}

To validate the results of the ALA method and assess the transportation system-level performance of integrating the Deep Activity model with ALA in a large transportation network, we conduct a series of experiments. As shown in Fig. \ref{fig:la_map} (a), the LA County region is divided into eight sub-regions. We use 100,000 agents from the SCAG ABM dataset as a training set to transfer the mobility model initially trained on the NHTS dataset, and fine-tune the ALA to obtain reference distributions of distances and angles for each sub-region. The fine-tuned mobility generation model and ALA are then applied to a population sample of 1 million to evaluate its validity and scalability.

\begin{figure}[ht]
  \centering
  \includegraphics[width=0.49\textwidth]{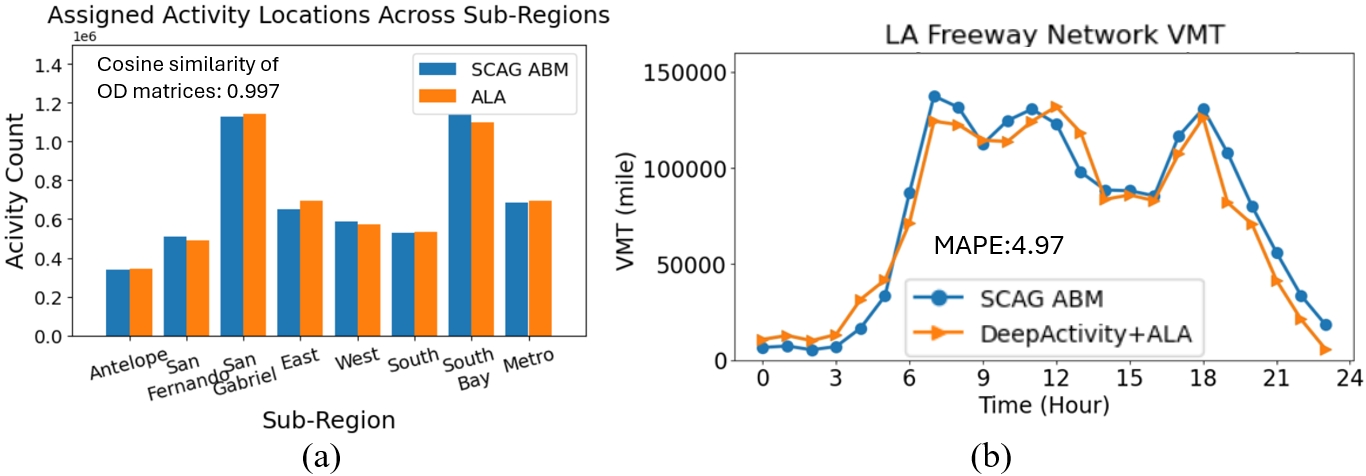}
  \caption{Validation for ALA and traffic loading at network level.}
  \label{fig:ala_network}
\end{figure}

Fig. \ref{fig:ala_network} (a) presents the distribution of activity locations across sub-regions in LA County. The results demonstrate that the ALA effectively captures the spatial distribution of activity locations across all sub-regions. The cosine similarity of OD matrices between SCAG ABM and ALA is 0.997, indicating the flow pattern generated by ALA  closely aligns with that of SCAG ABM. With the mobility patterns and assigned locations generated, we further load the travel demand into the LA transportation network, as depicted in Fig. \ref{fig:la_map} (b). The system-level traffic performance is illustrated in Fig. \ref{fig:ala_network} (b), which shows the hourly VMT and traffic speed over 24 hours, aggregated across all freeway segments. From a network-wide perspective, the proposed Deep Activity model and ALA collectively ensure a well-aligned temporal distribution of traffic flow, yielding MAPEs of 4.97 for VMT and 1.16 for traffic speed when compared to benchmark results from SCAG ABM, demonstrating strong system-level performance.

\begin{figure}[ht]
  \centering
  \includegraphics[width=0.49\textwidth]{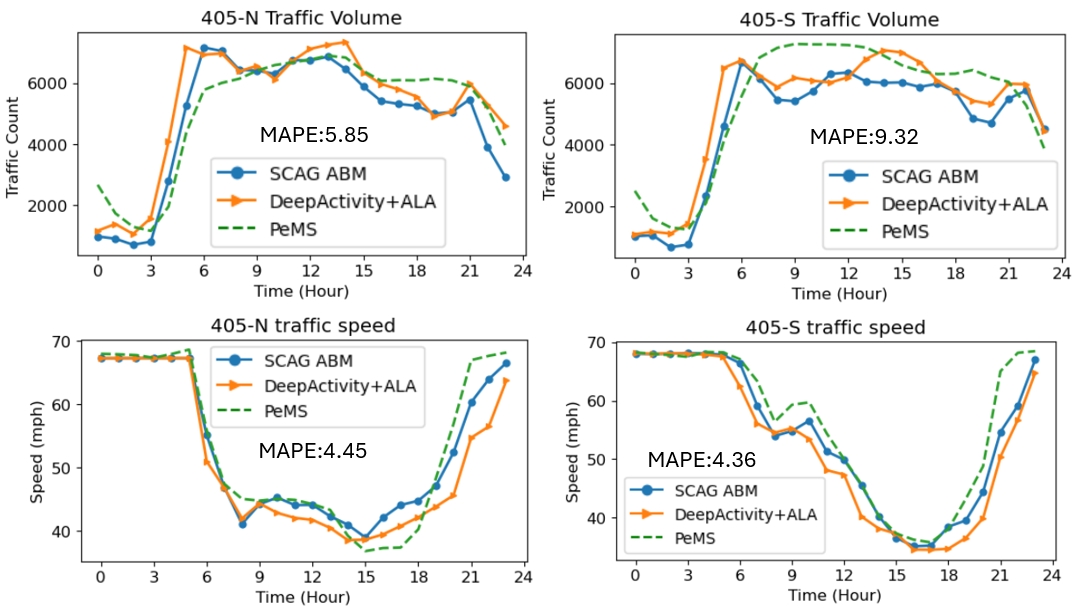}
  \caption{Validation for ALA and traffic loading at corridor level.}
  \label{fig:ala_corridor}
\end{figure}

Beyond the system-wide traffic metrics, we conduct further analysis at the corridor level by selecting a major segment from Interstate 405, a key freeway in the LA network. The location of the selected freeway segment is highlighted by the purple rectangle in Fig. \ref{fig:ala_network} (b). The 24-hour traffic volume and speed in both directions on the target corridor segment are shown in Fig. \ref{fig:ala_corridor} for comparison. To ensure the fidelity of the simulation, we include real-world observation data from the Caltrans Performance Measurement System (PeMS) \cite{pems2020} as a reference and compare the results from our proposed model with those from the SCAG ABM data. Note that the MAPE is just calculated between the proposed model results and SCAG ABM, as PeMS data serve only as a reference.
As seen in Fig. \ref{fig:ala_corridor}, PeMS observations indicate that both directions of the selected corridor experience high traffic volumes and significant congestion during the daytime. Notably, congestion patterns differ between the two directions: the northbound direction experiences major congestion throughout most of the midday, from 6 AM to 8 PM, while the southbound direction's primary congestion occurs after 12 PM and continues into the evening. The MAPE for traffic volume in the northbound and southbound directions is 5.85 and 9.32, respectively, while the MAPE for traffic speed in the northbound and southbound directions is 4.45 and 4.36, respectively.

These comparisons suggest that the proposed Deep Activity model and ALA successfully capture the dynamic temporal variations in traffic flow at the corridor level, demonstrating the model's good representation of the transportation system from both a travel demand generation and traffic loading perspective.

\subsection{The Influence of Model Complexity and Data Size}\label{section:dataComplexity}
The Deep Activity model was trained on the NHTS dataset, which has a modest sample size, potentially limiting the training effectiveness of complex models like Transformers. In this section, we explore the relationship between model complexity and data size in activity generation tasks.

To determine the optimal balance between model complexity and dataset size, we evaluated nine Transformer configurations on the NHTS dataset (180,000 samples). These models, which varied in the number of decoder layers (D), encoder layers (E), and attention heads (H), represented a spectrum of complexity, with parameters ranging from 434,000 to 2,344,000.

\begin{figure}
  \centering
  \includegraphics[width=0.49\textwidth]{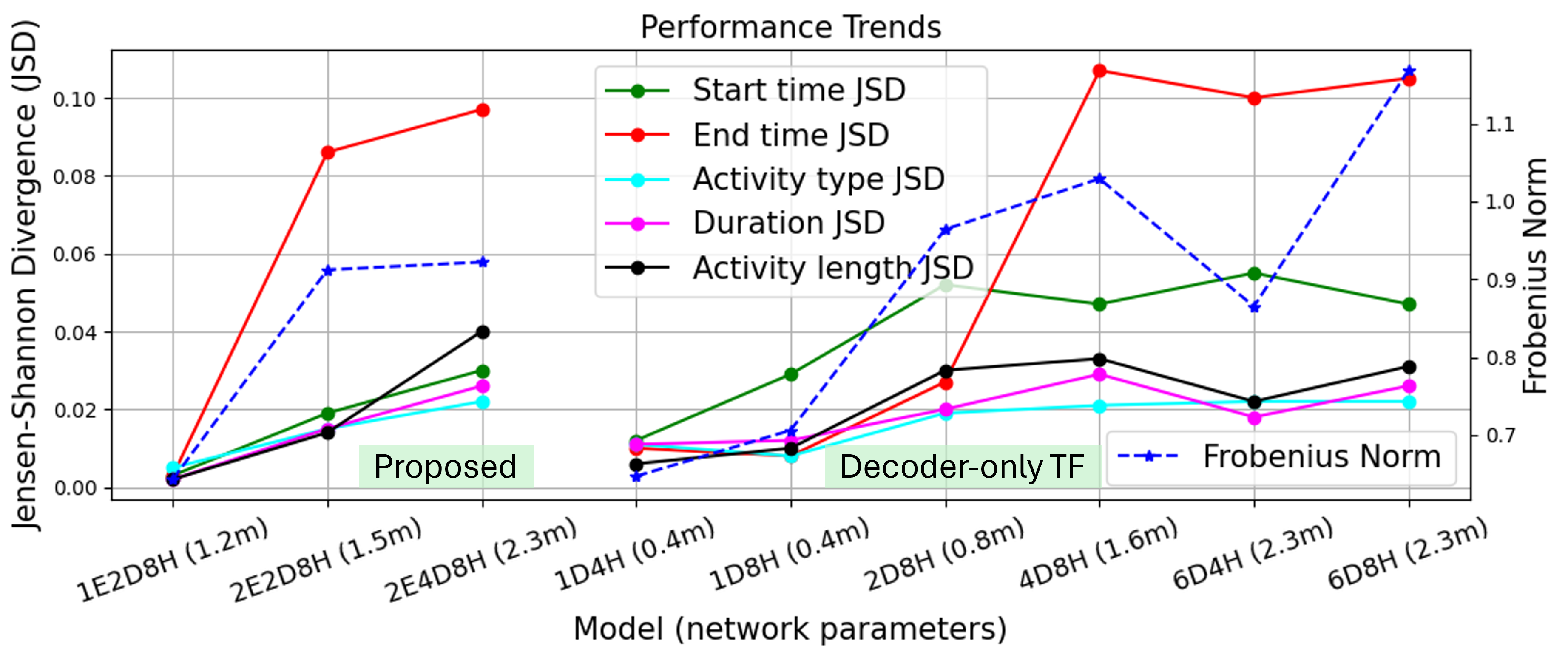}
  \caption{Performance evaluation for transformer models with different complexity.}
  \label{fig:ModelComplexity}
  \vspace{-2mm}

\end{figure}

Our analysis, as illustrated in Fig. \ref{fig:ModelComplexity}, indicated that models with fewer layers generally performed better, as reflected by lower JSD values and Frobenius norms. Simpler decoder-only models and those with a balanced encoder-decoder structure showed competitive results. Moreover, increasing the number of attention heads had only a moderate effect on performance. This suggests that, for the given dataset size, adding more layers or attention heads does not guarantee improved performance and may lead to diminishing returns or increased complexity without significant accuracy gains.

\begin{figure}
  \centering
  \includegraphics[width=0.49\textwidth]{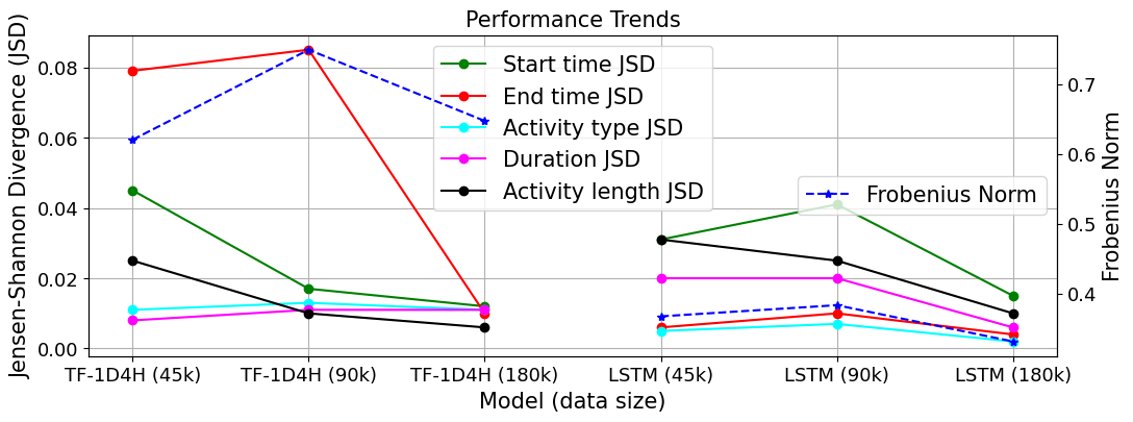}
  \caption{Training data size effect on decoder-only transformer and LSTM.}
  \label{fig:DataModelPerformance}
  \vspace{-2mm}

\end{figure}

To further understand the scaling effects of data size on different models, we compared a Transformer model with an LSTM model, as shown in Fig. \ref{fig:DataModelPerformance}. The Transformer exhibited improved performance with increased data size, from 45,000 to 180,000 samples, whereas the LSTM model's performance plateaued, indicating its limited ability to benefit from larger datasets. These results highlight the Transformer's superior capacity for utilizing larger datasets, aided by its parallel processing capabilities and global receptive field.

\textbf{Model Selection.} Given the modest sample size of the NHTS dataset, our findings suggest that: \textbf{1)} training on NHTS dataset, Transformer-based models with fewer layers perform better; \textbf{2)} the number of attention heads has moderate influence on the performance; \textbf{3)} Transformer-based model leverage larger datasets better, with performance increased when data size increases; \textbf{4)} LSTM-based models reach their limitation in current dataset and plateau with increased data size. Moreover, the explainability (as in Section \ref{sec:explanability}) and transferability (as in Section \ref{Sec:transfer}) of transformer-based models resulted in choosing the transformer-based Deep Activity model.

\section{Conclusion and Future Work}\label{sec:conclusion}
In this paper, we proposed the Deep Activity Model, a generative deep learning approach for human mobility synthesis that serves as a prototype and a foundational step toward high-fidelity activity generation. We adopt the concept of "activity chains" to accurately represent the daily mobility patterns of individuals by applying household travel survey data in deep learning to model human mobility patterns, showcasing a pioneering method in the field. The Deep Activity model effectively generates realistic activity chains with high fidelity through a robust location assignment algorithm. Our experiments demonstrated the model's versatility and robustness, achieving strong performance across various regions, including California, the Puget Sound area, and Mexico City. These results underscore the model's adaptability in generating diverse human mobility patterns, highlighting its potential for broader applications in urban planning and transportation analysis while maintaining a focus on data privacy.

The Deep Activity model serves as the foundation model, there are opportunities for future work. In particular, for applications that require highly accurate modeling, the adaptive temporal loss weighting and activity-type-specific temporal priors to improve performance in underrepresented time windows—especially. On the other hand, while the NHTS is a widely used and comprehensive dataset, it may exhibit sampling or reporting biases that underrepresent certain population segments or rare activity patterns. These biases can lead to disproportionate learning and limit the model's generalizability. In addition, our current approach is constrained by the limitations of existing household travel survey datasets, which often lack precise location information. As a result, our location assignment is currently restricted to the traffic analysis zone level. Incorporating more detailed location data could improve spatial accuracy and broaden the model's applicability. However, existing GPS datasets often lack semantic context, such as activity type. Addressing this gap through human trajectory data mining, as explored in our other study \cite{liu2024semantic}, could help associate semantic information with GPS traces. Integrating point of interest level location data with detailed activity and socio-demographic information would support more comprehensive modeling and enable deeper insights into human mobility patterns. Additionally, while our model demonstrates strong performance with Transformer-based architecture, we acknowledge the potential of alternative generative approaches such as conditional variational autoencoders (CVAEs), generative adversarial networks (GANs), and diffusion models. These architectures show promises for human mobility synthesis but often require dense, continuous mobility traces, larger datasets, and distinct training objectives. If suitable datasets become available, incorporating such models represents a valuable direction for future research. We plan to explore these architectures in a dedicated follow-up study. These advancements would support more advanced applications in urban planning, transportation management, and policy-making.

\section*{Acknowledgments}
This study is supported by the Intelligence Advanced Research Projects Activity (IARPA) via Department of Interior/Interior Business Center (DOI/IBC) contract number 140D0423C0033. The U.S. Government is authorized to reproduce and distribute reprints for Governmental purposes notwithstanding any copyright annotation thereon. Disclaimer: The views and conclusions contained herein are those of the authors and should not be interpreted as necessarily representing the official policies or endorsements, either expressed or implied, of IARPA, DOI/IBC, or the U.S. Government.

\bibliographystyle{IEEEtran}
\bibliography{reference}

\ifCLASSOPTIONcaptionsoff
  \newpage
\fi

\vskip 0pt plus -1fil
\begin{IEEEbiography}
[{\includegraphics[width=1in,height=1.25in,clip,keepaspectratio]{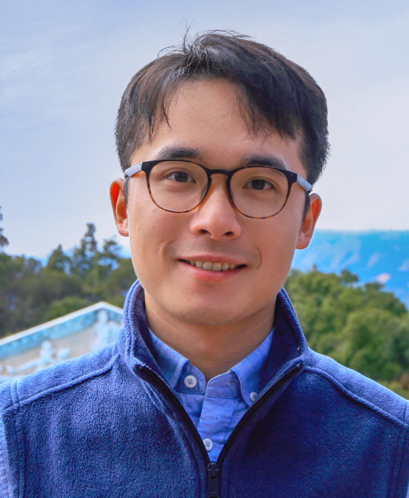}}]
{Xishun Liao}
received his Ph.D. degree in Electrical and Computer Engineering from the University of California, Riverside in 2023, the M.Eng. degree in Mechanical Engineering from the University of Maryland, College Park in 2018, and the B.E. degree in Mechanical Engineering and Automation from the Beijing University of Posts and Telecommunications in 2016. He was an assistant project scientist at the University of California, Los Angeles. Currently, he is an Assistant Professor at the University of Central Florida. His research focuses include smart mobility systems, human mobility modeling, urban geospatial intelligence, autonomous driving, applied AI/ML, and digital twin technology.
\end{IEEEbiography}

\vskip 0pt plus -1fil
\begin{IEEEbiography}
[{\includegraphics[width=1in,height=1.25in,clip,keepaspectratio]{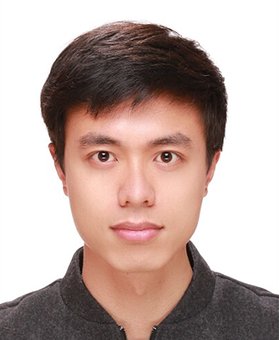}}]
{Qinhua Jiang} 
is a Ph.D. candidate from the Civil and Environmental Engineering Department, University of California, Los Angeles. He received his B.S. and M.S. degrees in civil engineering from Beijing Jiaotong University. His areas of expertise include activity-based travel demand modeling, AI/machine learning based traffic forecasting, and large-scale intelligent transportation system analysis.
\end{IEEEbiography}

\vskip 0pt plus -1fil
\begin{IEEEbiography}
[{\includegraphics[width=1in,height=1.25in,clip,keepaspectratio]{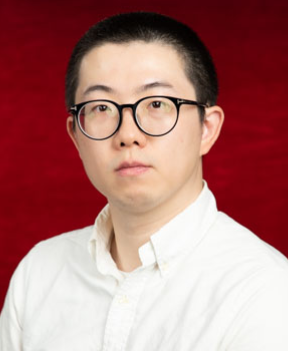}}]
{Yueshuai He} 
received his Ph.D. in Transportation Planning and Engineering from New York University in 2020. Prior to his current role as an assistant professor at the University of Louisville, he was a research scientist at UCLA. He has extensive research experience in transportation system modeling, human mobility study, and sustainable mobility system planning and management. His work focuses on developing advanced statistical and computational simulation models to understand transportation system dynamics and facilitate the decision-making of public agencies.
\end{IEEEbiography}

\vskip 0pt plus -1fil
\begin{IEEEbiography}
[{\includegraphics[width=1in,height=1.25in,clip,keepaspectratio]{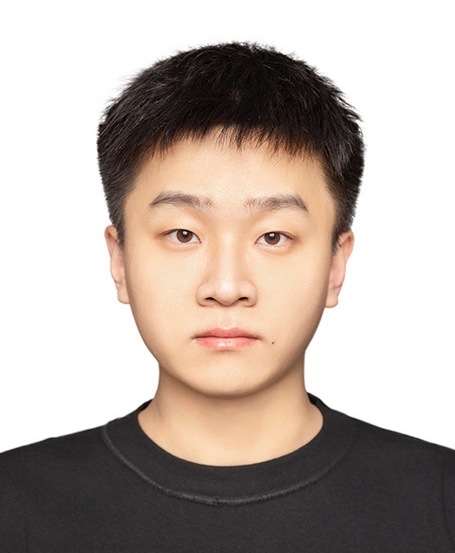}}]
{Yifan Liu} 
is a Ph.D. student from the Civil and Environmental Engineering Department, University of California, Los Angeles. He received his M.S. degree in Computer Science from New York University in 2024 and his B.E. degree in Computer Science and Engineering from the University of California, Davis. His research interests include large language models, reinforcement learning, and deep learning, with a focus on applications to smart mobility systems and autonomous driving.

\end{IEEEbiography}

\vskip 0pt plus -1fil
\begin{IEEEbiography}
[{\includegraphics[width=1in,height=1.25in,clip,keepaspectratio]{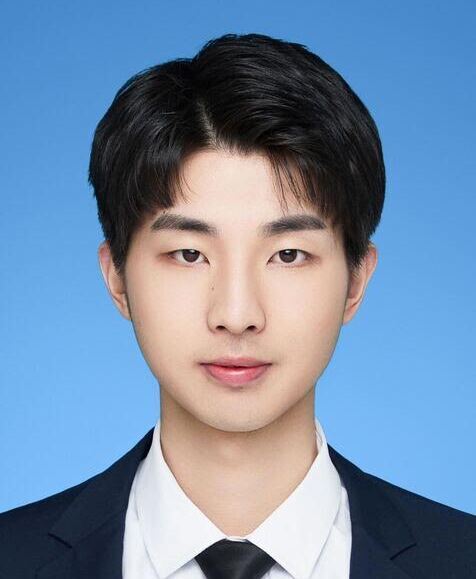}}]
{Chenchen Kuai} 
received his M.S. in Transportation Engineering from the University of California, Los Angeles in 2024, and his B.S. in Highway and Bridge Engineering from Southeast University in 2022. He is currently pursuing a Ph.D. degree in the Zachary Department of Civil and Environmental Engineering at Texas A\&M University. His research interests include Intelligent Transportation Systems, Complex Networks, Deep Learning, and Human Mobility.
\end{IEEEbiography}

\vskip 0pt plus -1fil
\begin{IEEEbiography}
[{\includegraphics[width=1in,height=1.25in,clip,keepaspectratio]{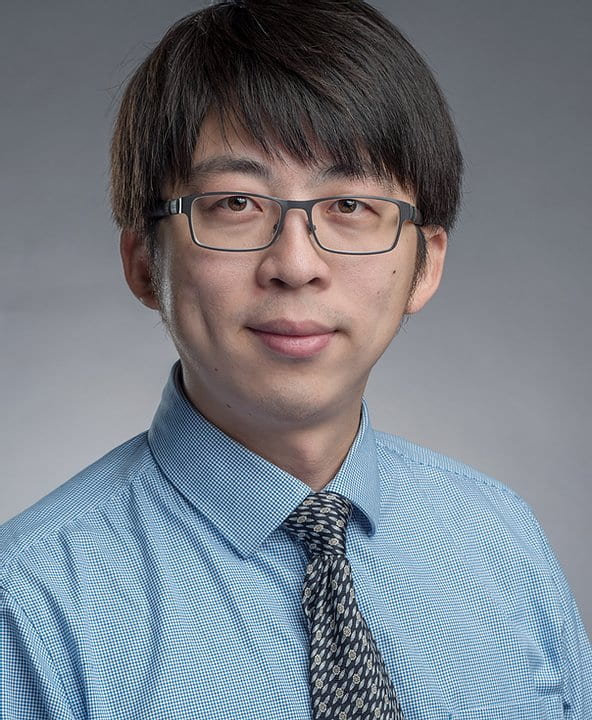}}]
{Jiaqi Ma} 
(Senior Member, IEEE) received the Ph.D. degree in transportation engineering from the University of Virginia, Charlottesville, VA, USA, in 2014. He is currently an Professor with the Samueli School of Engineering, UCLA, where he is also the Faculty Lead of the New Mobility Program, Institute of Transportation Studies. His research interests include intelligent transportation systems, autonomous driving, and cooperative driving automation. Dr. Ma serves as Editor-in-Chief of the IEEE Open Journal of Intelligent Transportation Systems, and Senior Editor of IEEE Transactions on Intelligent Transportation Systems. He is Chair of the Transportation Research Board (TRB) Standing Committee on Connected and Automated Vehicle Systems.

\end{IEEEbiography}

\end{document}